\documentclass[manuscript,screen,nonacm]{acmart}
\usepackage{tabularx,graphicx}
\usepackage{placeins}
\usepackage{tikz}
\usetikzlibrary{positioning,arrows.meta}
\newcommand{\R}{\mathbb{R}}
\newcommand{\ind}{\mathbf{1}}

\newcommand{\WithinGain}{0.167}
\newcommand{\WithinCI}{[0.123,\,0.214]}
\newcommand{\AblRemBase}{4.41}
\newcommand{\AblRemBaseCI}{[1.81, 7.28]}
\newcommand{\AblRemActive}{4.28}
\newcommand{\AblIncNorm}{$-$4.60}
\newcommand{\AblIncNormCI}{[$-$9.43, $-$0.62]}
\newcommand{\DonorTauLo}{$-$2.05}
\newcommand{\DonorTauHi}{$-$1.41}

\newcommand{\CtlEightDecisionRaw}{curvature-limited}

\newcommand{\CtlValidatedTests}{five}

\newcommand{\CtlPMAdmitted}{332}
\newcommand{\CtlPMChecks}{30}
\newcommand{\CtlPMFVUCentered}{0.14}
\newcommand{\CtlPMFVURaw}{1.01}
\newcommand{\CtlPMCap}{5.61}

\newcommand{\CtlPMThetaZRaw}{4.3}

\newcommand{\CtlPMBootLoRaw}{0.79}
\newcommand{\CtlPMBootHiRaw}{2.03}

\newcommand{\CtlPMThetaZSae}{4.3}

\newcommand{\CtlPMBootLoSae}{0.73}
\newcommand{\CtlPMBootHiSae}{1.88}

\newcommand{\CtlPMTwoAdmitted}{889}
\newcommand{\CtlPMTwoChecks}{38}
\newcommand{\CtlPMTwoFVU}{0.48}
\newcommand{\CtlPMTwoFVUResid}{1.16}
\newcommand{\CtlPMTwoCap}{5.17}
\newcommand{\CtlPMTwoVtwoZ}{1.1}

\newcommand{\CtlOtherAllMin}{1}
\newcommand{\CtlOtherAllMax}{46}

\newcommand{\CtlPMTwoVtwoInfMin}{4.0}
\newcommand{\CtlPMTwoVtwoInfMax}{4.5}
\newcommand{\CtlPremisePublicCount}{2,444}
\newcommand{\CtlPremisePublicBound}{$10^{-14}$}
\newcommand{\CtlPremiseCount}{2,034}
\newcommand{\CtlPremiseBound}{$10^{-14}$}

\newcommand{\CtlStarZMin}{2.2}
\newcommand{\CtlStarZMax}{3.9}

\newcommand{\CtlRatioMin}{1.10}
\newcommand{\CtlRatioMax}{1.51}

\newcommand{\CtlEightCap}{9.9}
\newcommand{\CtlThreeCap}{5.08}

\newcommand{\CtlOutRatioMin}{9.8}
\newcommand{\CtlOutRatioMax}{15.9}
\newcommand{\CtlEightNativeRaw}{0.71}
\newcommand{\CtlEightCrossedRaw}{8 of 9}

\newcommand{\CtlEightNativeSae}{0.75}
\newcommand{\CtlEightCrossedSae}{7 of 9}

\newcommand{\CtlPMThreeRegWrong}{14}
\newcommand{\CtlPyFVU}{0.28}
\newcommand{\CtlPyFVUWrong}{1.38}
\newcommand{\CtlPyCap}{2.71}
\newcommand{\CtlPyCapCost}{15}
\newcommand{\CtlPyGatePass}{5}
\newcommand{\CtlPyGateMiss}{1.28}
\newcommand{\CtlPyAdmittedFp}{928}
\newcommand{\CtlPyPredLo}{0.03}
\newcommand{\CtlPyPredHi}{0.87}
\newcommand{\CtlPyNoiseLo}{$-$1.28}
\newcommand{\CtlPyNoiseHi}{2.26}
\newcommand{\CtlPyDiagLo}{1.000}
\newcommand{\CtlPyDiagHi}{1.023}
\newcommand{\CtlPyDiagLast}{0.003}

\newcommand{\CtlPyDistMin}{1.06}
\newcommand{\CtlPyDistMax}{1.17}
\newcommand{\CtlAtwoRegFpWrong}{56}
\newcommand{\CtlAtwoRegDocs}{8}
\newcommand{\CtlGmFVU}{0.12}
\newcommand{\CtlGmFVUWrong}{0.21}
\newcommand{\CtlGmCap}{11.08}
\newcommand{\CtlGmCapCost}{65,103}
\newcommand{\CtlGmPromoted}{157}
\newcommand{\CtlGmAdmitted}{883}
\newcommand{\CtlGmNoop}{117}

\newcommand{\CtlGmGateLo}{0.997}
\newcommand{\CtlGmGateHi}{1.158}

\newcommand{\CtlLlFVU}{0.35}
\newcommand{\CtlLlFVUWrong}{1.08}
\newcommand{\CtlLlCap}{5.08}
\newcommand{\CtlLlCapCost}{161}
\newcommand{\CtlLlPromoted}{33}
\newcommand{\CtlLlAdmitted}{680}
\newcommand{\CtlLlNoop}{306}
\newcommand{\CtlLlNotCon}{14}
\newcommand{\CtlLlGateLo}{0.996}
\newcommand{\CtlLlGateHi}{1.000}

\newcommand{\CtlPubDeclN}{fourteen}
\newcommand{\CtlPubDeclYes}{thirteen}
\newcommand{\CtlDecProbeMax}{2.4\times10^{-6}}

\newcommand{\CtlPyBlindMax}{3.9}
\newcommand{\CtlPyOtherFactor}{2.4}

\newcommand{\CtlXsLlResidMax}{81}

\newcommand{\CtlSecXsHold}{four}
\newcommand{\CtlSecXsFail}{two}
\newcommand{\CtlSecXsDirections}{24}
\newcommand{\CtlSecXsRemovedMin}{95}
\newcommand{\CtlSecXsRemovedMax}{105}
\newcommand{\CtlSecXsResidMin}{$-$5}
\newcommand{\CtlSecXsResidMax}{5}
\newcommand{\CtlSecXsLeftMin}{0.9}
\newcommand{\CtlSecXsLeftMax}{3.8}
\newcommand{\CtlSecXsLeftZMin}{56}
\newcommand{\CtlSecXsLeftZMax}{474}
\newcommand{\CtlSecRepN}{twelve}
\newcommand{\CtlSecRepYes}{ten}

\newcommand{\CtlSecRepNoList}{Llama on TWOS, block 1, raw metric; Gemma on LANL, block 2, raw metric}
\newcommand{\CtlSecRepLeftMin}{1.9}
\newcommand{\CtlSecRepLeftMax}{2.2}
\newcommand{\CtlSecRepLeftZMin}{2.1}
\newcommand{\CtlSecRepLeftZMax}{3.0}
\newcommand{\CtlSecRepLeftThetaZMin}{74}
\newcommand{\CtlSecRepLeftThetaZMax}{89}
\newcommand{\CtlSecRepLeftClusterTMin}{36}
\newcommand{\CtlSecRepLeftClusterTMax}{51}
\newcommand{\CtlSecRepLeftRatioMin}{0.978}
\newcommand{\CtlSecRepLeftRatioMax}{0.981}

\newcommand{\CtlSecOtherMax}{46}
\newcommand{\CtlSecRatioMin}{0.98}
\newcommand{\CtlSecRatioMax}{1.01}
\newcommand{\CtlSecGMin}{5.9}
\newcommand{\CtlSecGMax}{11.5}
\newcommand{\CtlSecFVUMin}{0.25}
\newcommand{\CtlSecFVUMax}{0.55}
\newcommand{\CtlSecFVUWrongMin}{0.47}
\newcommand{\CtlSecFVUWrongMax}{1.41}
\newcommand{\CtlSecAdmittedMin}{910}
\newcommand{\CtlSecAdmittedMax}{1,000}
\newcommand{\CtlSecGateLo}{0.995}
\newcommand{\CtlSecGateHi}{1.007}
\newcommand{\CtlSecDocsTwosMin}{4,414}

\newcommand{\CtlSecDocsLanlMin}{9,883}
\newcommand{\CtlSecDocsLanlMax}{9,918}
\newcommand{\CtlXFivePstarRaw}{31}

\newcommand{\CtlXFiveRandomRaw}{9}
\newcommand{\CtlXFiveRatioRaw}{1.72}
\newcommand{\CtlXFivePstarSae}{30}

\newcommand{\CtlXFiveRandomSae}{17}
\newcommand{\CtlXFiveRatioSae}{0.99}
\newcommand{\CtlXFiveAdmitted}{67}
\newcommand{\CtlXFiveVerdict}{does not replicate}
\newcommand{\CtlXFiveXFourVerdict}{uninformative}
\newcommand{\CtlXFiveHOneRaw}{4}
\newcommand{\CtlXFiveHOneSae}{3}

\newcommand{\CtlXFiveFortyFourPstarRaw}{26}

\newcommand{\CtlXFiveFortyFourRandomRaw}{5}
\newcommand{\CtlXFiveFortyFourRatioRaw}{1.06}
\newcommand{\CtlXFiveFortyFourPstarSae}{27}

\newcommand{\CtlXFiveFortyFourRandomSae}{6}
\newcommand{\CtlXFiveFortyFourRatioSae}{1.04}
\newcommand{\CtlXFiveFortyFourAdmitted}{73}
\newcommand{\CtlXFiveFortyFourVerdict}{replicates}
\newcommand{\CtlXFiveFortyFourXFourVerdict}{supported}
\newcommand{\CtlXFiveFortyFourHOneRaw}{13}
\newcommand{\CtlXFiveFortyFourHOneSae}{11}
\newcommand{\CtlXFiveFortyFourHOneMedRaw}{0.17}
\newcommand{\CtlXFiveFortyFourHOneMedSae}{0.15}
\newcommand{\CtlXFiveFortyFivePstarRaw}{28}

\newcommand{\CtlXFiveFortyFiveRandomRaw}{7}
\newcommand{\CtlXFiveFortyFiveRatioRaw}{1.12}
\newcommand{\CtlXFiveFortyFivePstarSae}{25}

\newcommand{\CtlXFiveFortyFiveRandomSae}{12}
\newcommand{\CtlXFiveFortyFiveRatioSae}{0.90}
\newcommand{\CtlXFiveFortyFiveAdmitted}{64}
\newcommand{\CtlXFiveFortyFiveVerdict}{does not replicate}
\newcommand{\CtlXFiveFortyFiveXFourVerdict}{uninformative}
\newcommand{\CtlXFiveFortyFiveHOneRaw}{8}
\newcommand{\CtlXFiveFortyFiveHOneSae}{9}
\newcommand{\CtlXFiveFortyFiveHOneMedRaw}{0.09}
\newcommand{\CtlXFiveFortyFiveHOneMedSae}{0.08}
\newcommand{\CtlXFiveFalsifRaw}{0.62}
\newcommand{\CtlXFiveSeedsN}{three}
\newcommand{\CtlXFiveSeedsRep}{one}
\newcommand{\CtlXFiveAdmMin}{64}
\newcommand{\CtlXFiveAdmMax}{73}
\newcommand{\CtlXFiveUninfMin}{3}
\newcommand{\CtlXFiveUninfMax}{9}

\newcommand{\CtlXSixGemmaAdmitted}{89}
\newcommand{\CtlXSixGemmaXFourVerdict}{uninformative}
\newcommand{\CtlXSixGemmaHOneRaw}{1}
\newcommand{\CtlXSixGemmaHOneSae}{2}
\newcommand{\CtlXSixGemmaPstarRaw}{43}
\newcommand{\CtlXSixGemmaThetaRaw}{$-$0.10}
\newcommand{\CtlXSixGemmaGapRaw}{74}
\newcommand{\CtlXSixGemmaRemovedRaw}{4}

\newcommand{\CtlXSixGemmaFalsifRaw}{0.62}

\newcommand{\CtlXSixGemmaPstarSae}{42}
\newcommand{\CtlXSixGemmaThetaSae}{$-$0.10}
\newcommand{\CtlXSixGemmaGapSae}{93}
\newcommand{\CtlXSixGemmaRemovedSae}{61}

\newcommand{\CtlXSixGemmaFalsifSae}{0.60}

\newcommand{\CtlXSixGemmaVerdict}{does not replicate}
\newcommand{\CtlXSixGemmaKThreeRaw}{3.54}
\newcommand{\CtlXSixGemmaKThreeSae}{7.17}
\newcommand{\CtlXSixGemmaTInfRaw}{1}
\newcommand{\CtlXSixGemmaTInfSae}{2}
\newcommand{\CtlXSixGemmaTOf}{14}
\newcommand{\CtlXSixLlamaAdmitted}{3}
\newcommand{\CtlXSixLlamaXFourVerdict}{uninformative}

\newcommand{\CtlXSixLlamaPsample}{3}

\newcommand{\CtlXSixLlamaNotConstructed}{597}

\newcommand{\CtlXFiveOf}{600}
\newcommand{\CtlXFiveSeedFortyTwo}{196}
\newcommand{\CtlNTwoTwosDin}{11}
\newcommand{\CtlNTwoLanlDin}{33}
\newcommand{\CtlNTwoTwosPMin}{75}
\newcommand{\CtlNTwoTwosPMax}{93}
\newcommand{\CtlNTwoLanlGMax}{2.1}
\newcommand{\CtlNTwoTwosDeffNarrow}{33}
\newcommand{\CtlNTwoTwosRandNarrow}{6.8}
\newcommand{\CtlNTwoTwosAucNarrow}{0.76}
\newcommand{\CtlNTwoTwosDeffWide}{402}
\newcommand{\CtlNTwoTwosRandWide}{10.2}
\newcommand{\CtlNTwoTwosAucWide}{0.60}
\newcommand{\CtlNTwoLanlDeffNarrow}{23}
\newcommand{\CtlNTwoLanlRandNarrow}{17.9}
\newcommand{\CtlNTwoLanlAucNarrow}{0.54}
\newcommand{\CtlNTwoLanlDeffWide}{193}
\newcommand{\CtlNTwoLanlRandWide}{12.4}
\newcommand{\CtlNTwoLanlAucWide}{0.56}
\newcommand{\CtlNTwoLanlRandMax}{55}
\newcommand{\CtlWebRepEighty}{11}
\newcommand{\CtlWebRepEightyN}{12}
\newcommand{\CtlWebRepGemmaMiss}{0.10}
\newcommand{\CtlWebRepGemmaLeft}{$-$7}
\newcommand{\CtlXSevenTwosAAdmitted}{359}
\newcommand{\CtlXSevenTwosAXFourVerdict}{supported}

\newcommand{\CtlXSevenTwosAHOneRaw}{89}
\newcommand{\CtlXSevenTwosAHOneMedRaw}{0.27}

\newcommand{\CtlXSevenTwosATInfRaw}{11}
\newcommand{\CtlXSevenTwosAHOneSae}{95}
\newcommand{\CtlXSevenTwosAHOneMedSae}{0.28}

\newcommand{\CtlXSevenTwosATInfSae}{9}
\newcommand{\CtlXSevenTwosATOf}{11}
\newcommand{\CtlXSevenTwosBAdmitted}{76}

\newcommand{\CtlXSevenTwosBHOneRaw}{49}
\newcommand{\CtlXSevenTwosBHOneMedRaw}{0.08}

\newcommand{\CtlXSevenTwosBTInfRaw}{11}
\newcommand{\CtlXSevenTwosBHOneSae}{47}
\newcommand{\CtlXSevenTwosBHOneMedSae}{0.08}

\newcommand{\CtlXSevenTwosBTInfSae}{11}
\newcommand{\CtlXSevenTwosBTOf}{11}

\newcommand{\CtlXSevenTwosANoop}{235}
\newcommand{\CtlXSevenTwosANotConstructed}{6}
\newcommand{\CtlXSevenTwosBNoop}{523}
\newcommand{\CtlXSevenTwosBNotConstructed}{1}
\newcommand{\CtlXEightPstarRaw}{143}
\newcommand{\CtlXEightTInfRaw}{33}
\newcommand{\CtlXEightLeftRaw}{$-$9}
\newcommand{\CtlXEightThetaTRaw}{$-$1.4}
\newcommand{\CtlXEightRandomRaw}{17}
\newcommand{\CtlXEightPstarSae}{145}
\newcommand{\CtlXEightTInfSae}{32}
\newcommand{\CtlXEightLeftSae}{$-$8}
\newcommand{\CtlXEightThetaTSae}{$-$0.8}
\newcommand{\CtlXEightRandomSae}{27}
\newcommand{\CtlXEightTOf}{46}
\newcommand{\CtlXEightFalsifRaw}{0.78}
\newcommand{\CtlXEightAdmitted}{240}
\newcommand{\CtlXEightAdmittedUsers}{154}
\newcommand{\CtlXEightNotConstructed}{354}
\newcommand{\CtlXEightNoop}{6}
\newcommand{\CtlXEightXFourVerdict}{uninformative}
\newcommand{\CtlXEightRCap}{4.70}
\newcommand{\CtlXEightSignifRaw}{130}
\newcommand{\CtlXEightKOneRaw}{1.32}

\newcommand{\CtlXEightSignifSae}{134}
\newcommand{\CtlXEightKOneSae}{1.21}

\newcommand{\CtlSetRuns}{34}
\newcommand{\CtlSetModels}{seven}
\newcommand{\CtlSetRunsCore}{33}

\newcommand{\CtlSetRemovedMin}{58}
\newcommand{\CtlSetGapMax}{18}
\newcommand{\CtlSetCostMin}{1.06}
\newcommand{\CtlSetCostMax}{6.54}
\newcommand{\CtlSetBlindMax}{27.5}
\newcommand{\CtlSetBlindMaxNoSeed}{11.0}

\newcommand{\CtlResLeftPairs}{18}
\newcommand{\CtlSetPoints}{68}
\newcommand{\CtlSetPrimary}{15}
\newcommand{\CtlSetReplications}{12}
\newcommand{\CtlAeGapMin}{17}
\newcommand{\CtlAeGapMax}{73}
\newcommand{\CtlAeBlindMax}{35}

\newcommand{\CtlResWebPairs}{10}
\newcommand{\CtlResWebHold}{8}

\newcommand{\CtlResAePairs}{4}
\newcommand{\CtlResAeHold}{3}

\newcommand{\CtlUnitZWebMin}{4.3}
\newcommand{\CtlUnitZWebMax}{17.0}
\newcommand{\CtlUnitZLogsMin}{24.2}
\newcommand{\CtlUnitZLogsMax}{112.1}
\newcommand{\CtlUnitZCertMin}{67.9}
\newcommand{\CtlUnitZCertMax}{415.4}
\newcommand{\CtlResHold}{52}

\newcommand{\CtlResThreeThetaTMin}{2.2}
\newcommand{\CtlResThreeThetaTMax}{2.7}
\newcommand{\CtlResRandLow}{77}
\newcommand{\CtlResXSixGemmaThetaTRaw}{$-$2.2}
\newcommand{\CtlResXSixGemmaThetaTSae}{$-$2.1}

\newcommand{\CtlResXSixGemmaRandLow}{$-$21}
\newcommand{\CtlPMFivePyVerdict}{supports}

\newcommand{\CtlPMFiveGmVerdict}{supports}

\newcommand{\CtlAblRuns}{35}
\newcommand{\CtlAblPairs}{70}
\newcommand{\CtlAblDraws}{20,000}
\newcommand{\CtlAblNatN}{124}
\newcommand{\CtlAblNatOut}{4}
\newcommand{\CtlAblNatExp}{6}
\newcommand{\CtlWebRepRandMax}{16.7}
\newcommand{\CtlWebRepRandQ}{12.3}
\newcommand{\CtlAblNatMed}{0.62}
\newcommand{\CtlAblNatMedExp}{0.67}

\newcommand{\CtlAblLmLtwoMin}{6.5}
\newcommand{\CtlAblLmLtwoMax}{141}
\newcommand{\CtlAblLtwoOver}{three}
\newcommand{\CtlAblLmPathMin}{$-$5.0}
\newcommand{\CtlAblLmPathMax}{$+$7.8}
\newcommand{\CtlAblLmDeffMin}{199}
\newcommand{\CtlAblLmDeffMax}{1,804}
\newcommand{\CtlAblLmSdMax}{7.1}
\newcommand{\CtlAblAeDeffMin}{17}
\newcommand{\CtlAblAeDeffMax}{27}

\newcommand{\CtlAblQMaxTwo}{2.24}
\newcommand{\CtlWidthLoDeff}{24}
\newcommand{\CtlWidthLoRand}{16.7}
\newcommand{\CtlWidthMidDeff}{97}
\newcommand{\CtlWidthMidRand}{13.8}
\newcommand{\CtlWidthHiDeff}{326}
\newcommand{\CtlWidthHiRand}{4.0}
\newcommand{\CtlWidthThreePairs}{22}
\newcommand{\CtlWidthThreeFail}{one}

\newcommand{\CtlKsActMin}{$-$16}
\newcommand{\CtlKsActMax}{$-$35}
\newcommand{\CtlKsAttrThreeMin}{5}
\newcommand{\CtlKsAttrThreeMax}{7}
\newcommand{\CtlKsAttrFiveMin}{11}
\newcommand{\CtlKsAttrFiveMax}{19}
\newcommand{\CtlKsAttrTenMin}{32}
\newcommand{\CtlKsAttrTenMax}{61}

\setcopyright{none}
\renewcommand\footnotetextcopyrightpermission[1]{}
\begin{document}
\title{How Reliable Are Explanations of Language Models That Detect Insider Threats?}
\author{Suramya R. Angdembay}
\email{Suramya.Angdembay@usm.edu}
\affiliation{\institution{The University of Southern Mississippi}\city{Hattiesburg}\state{Mississippi}\country{USA}}

\begin{abstract}
Organizations monitor their workers' activity for insider threats, in which workers misuse their legitimate access,
by copying confidential files before leaving. Detectors score how unusual each day of activity is, flagging the most
unusual for an analyst to review. Prominent current approaches use opaque neural networks, so explanations of a flag
point to internal features: patterns of neuron activity most active on days of misuse. We audit such explanations
for language-model detectors trained on CERT's simulated workplace logs. To test an explanation, we see what happens
to the detector's score when we remove its features inside the model. We compare this to a control: a change that is
as similar as possible to the removal, but without actually removing the features. Because defining ``similar''
requires choosing a metric, our central contribution is the tipping radius: a first-order prediction of how far that metric must move before the
difference between removal and control disappears, a measure of how fragile the result is to that choice. In
\CtlSetRunsCore{} of \CtlSetRuns{} runs of \CtlSetModels{} language models, the tipping radius is at most
\CtlSetGapMax\% of the distance between two standard choices of metric. Moving that far removes at least
\CtlSetRemovedMin\% of the difference or reverses it, while random moves of the same size leave it largely intact. The
result replicates in \CtlPubDeclYes{} of \CtlPubDeclN{} pre-declared primary runs on openly released models, though on our
trained detectors the difference is often too small to measure. We also find that a flag's score partly reflects who
the worker is, not what they did, and that the features named depend on whom they are compared against. Audits should
therefore fix the control's metric in advance, report its tipping radius, and disclose that comparison.
\end{abstract}


\maketitle

\section{Introduction}
Employees need access to an organization's files, accounts and systems to do their jobs, and that access can be
misused, for instance to copy confidential files before leaving. Insider-threat detection looks for such misuse in
activity logs \citep{yuan2021deep}. Many detectors learn what ordinary activity looks like and give each day an
anomaly score; the most unusual days become alerts for an analyst to review \citep{chandola2009anomaly}. An alert is
not evidence of misconduct. A high score can arise because a day is unusual for its worker, because the worker differs
from the people the detector learned from, or because the record contains fields that are hard to predict. The score
alone does not tell these apart, so what an alert rests on matters to the workers being monitored as well as to the
organization (Section~\ref{sec:accountability}).

Prominent current approaches build detectors from neural networks, including language models
\citep{tuor2018recurrent,yuan2021deep,guo2021logbert,kantchelian2024facade,li2025redchronos,kong2025dmfi}. In a
language-model detector, each day of activity is written as text, the model is fine-tuned on the texts of workers with
no recorded misuse, and a day's anomaly score is how poorly the model predicts its text. The score shows which parts of
the text were hard to predict, but not what the model relied on. Explanations of an alert therefore look inside the
model \citep{cunningham2023sparse,gao2024scaling}: a sparse autoencoder (SAE) rewrites the model's internal state at
each position of the text as a combination of a few directions, called features, each a pattern in the activations of
the model's neurons. The features more active on days of misuse than on other workers' days are offered as the
explanation, and they invite strong readings, for instance that a feature encodes insider behavior. Whether those
features matter is tested by intervention: they are removed inside the model, and the change in the score is compared
with that under a control, an edit that should leave the explanation intact \citep{zhang2023best,heimersheim2024use}.

Each of these steps rests on a choice: which parts of the record the score counts, which workers the features are
compared against, and how the control is built. Prior work shows that such choices shape interpretability claims. In
activation patching, the metric that scores a patch and the way activations are corrupted can change which components
appear important \citep{zhang2023best,heimersheim2024use}; the value used to ablate a component changes its measured
importance \citep{li2024optimal}; common evaluations of neuron explanations fail basic sanity checks
\citep{oikarinen2025evaluating}; and choices like these can be searched for a preferred explanation
\citep{sharma2025xhacking} in ways that are hard to detect from outputs \citep{hinns2026cherry}.

One choice remains even when the features, the removal and the allowed controls are all fixed. The control is meant to
isolate the features' effect, so it is chosen as the edit most similar to the removal, and ``most similar'' requires a
measure of similarity. Two reasonable measures pick two different controls, and so two different measured effects,
for the same removal (Figure~\ref{fig:overview}, left). Rather than declaring one measure correct, we ask how sensitive the
conclusion is to that choice. This is our central question: how much do conclusions about a model's features depend on
the choices used to test them, and how far must the measure of similarity move before the conclusion changes?

\begin{figure}[t]
\centering
\includegraphics[width=\linewidth]{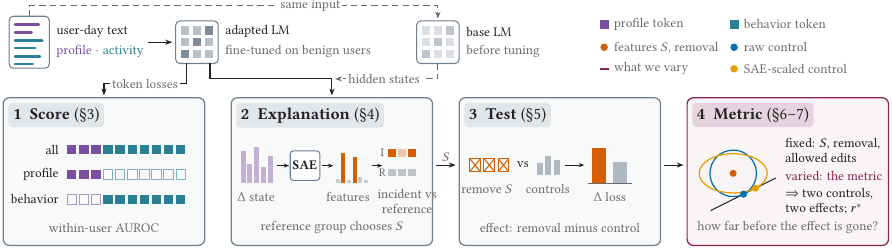}
\caption{The audit in four steps. The model's token losses feed step 1 and its hidden states step 2, so the explanation
does not start from the score's result; Figure~\ref{fig:overview} (Section~\ref{sec:geo-choice}) expands step 4.}
\label{fig:roadmap}
\end{figure}

We audit such explanations for language-model detectors trained on CERT, a simulated benchmark of an organization's
activity logs with scripted insider incidents \citep{certdataset,glasser2013bridging}. CERT calls workers users, as we do from here on, and calls the days of
an insider's scripted misuse incident days. Our principal detector is Qwen3-8B \citep{qwen3} (the 8B); supplementary
detectors use Qwen2.5-3B (the 3B). Sections~\ref{sec:score}--\ref{sec:interventions} show where an explanation depends
on these choices (Figure~\ref{fig:roadmap}): we ask \emph{what the score measures} (Section~\ref{sec:score}), \emph{how the explanation's features
are chosen} (Section~\ref{sec:sae}), and \emph{what removing them inside the model shows}
(Section~\ref{sec:interventions}). Section~\ref{sec:geometry} then measures \emph{how much that test depends on the
metric that defines its control}, and Section~\ref{sec:results}
tests that measurement across models and data.

The last step is our central contribution. The measure of similarity is a metric, a rule for how much a change in
each neuron's activation counts; the two standard metrics count every neuron equally (raw) or each relative to its
typical size (SAE-scaled). We introduce the \emph{tipping radius}: the smallest change to the control's metric that,
to first order, erases the explained effect, reported as a share of the distance between the two standard metrics. Like a
sensitivity parameter for hidden bias in an observational study \citep{vanderweele2017evalue},
it states how strong an unfavorable choice must be to change the conclusion.

In summary, we contribute:
\begin{itemize}
 \item an empirical audit of explanations of language-model insider-threat detectors, showing how the score, the
 choice of features and the test of those features each limit what an explanation can support
 (Sections~\ref{sec:score}--\ref{sec:interventions});
 \item the tipping radius, a sensitivity measure for the metric that defines a feature-preserving control, characterized to
 first order (Section~\ref{sec:geometry}) and tested against native reruns and same-size random changes in \CtlSetRuns{} runs of \CtlSetModels{} language models from five families, all but two declared before any
 outcome was computed (Section~\ref{sec:results}).
\end{itemize}
Together, the findings call for reporting how an explanation was tested, not only what it found
(Section~\ref{sec:implications}).
Our measurements concern prediction losses, internal edits and their controls. They do not show that any worker acted
maliciously, that an organization should deploy such a detector, or that our audit improves analysts' decisions;
Section~\ref{sec:related} sets out these limits. We do not propose a new SAE architecture or claim that language models
detect insiders better than other methods. Appendix~\ref{app:scope} records the provenance of each result.

\section{Background and related work}
\label{sec:accountability}
The security objective of insider-threat analysis can coexist with the interests of the workers whose activity is
scrutinized. CERT's incident-response guidance separates detecting an anomaly
from an inquiry intended to substantiate or refute it, and emphasizes confidentiality and documented response
procedures \citep{cert2015response}. This distinction motivates our audit: what does the evidence underlying an
alert establish before a person interprets it as a security concern?

Our experiments cover scoring and selected internal computations, within a sequence that runs from activity records
through alerts to inquiry and response. The workers in the records, the analysts reviewing alerts, the organizations
protecting assets and the auditors examining model claims have interests that one ranking metric cannot capture.

Profile information may provide legitimate predictive context. Its influence is not automatically discriminatory,
and excluding profile targets from the loss does not remove profile information from the input. The simulated
incident labels also do not make every action on an incident day malicious. Consequently, an explanation of
prediction error is not a finding about a person's intent. Access to model internals enables scrutiny
\citep{casper2024blackbox}; it does not itself resolve these interpretive limits. Even a faithful explanation would
leave the justification, proportionality and governance of a monitoring system to a broader assessment.

An explanation of an alert can be supported at three levels. A feature can be \emph{associated} with incidents, more
active on incident days than elsewhere; the model can \emph{depend} on it, so that changing the feature changes the
model's predictions; and the explanation is \emph{supported} only if that dependence is specific to the explained
behavior and survives the comparison and assumptions declared for it. None of the three shows that a feature represents
malicious intent. Sections~\ref{sec:sae}--\ref{sec:geometry} take these levels in turn,
which makes explicit the functionality question emphasized by \citet{raji2022functionality}.

\paragraph{Testing interpretability claims.} SAEs offer sparse coordinates for studying model states
\citep{cunningham2023sparse,gao2024scaling}. Beyond the patching, ablation and evaluation choices of the Introduction,
mediation scores for single inputs vary widely \citep{meloux2025variance} and small input perturbations can move SAE
concept readings \citep{li2026evaluating}. Causal claims in mechanistic interpretability rest on identification assumptions that
should be disclosed and stress-tested \citep{lin2026position}. These studies vary how an intervention is made or scored; we
hold the removal and the allowed controls fixed and vary only the metric that selects the control. Concept erasure
under a moment constraint has the same optimum under every inner-product norm \citep{belrose2023leace}; a control that
must reproduce fixed per-token readings does not, which is why its result changes with the metric.

\paragraph{Sensitivity analysis.} Like an E-value \citep{vanderweele2017evalue}, our tipping radius measures one
assumption, the control's metric, not all of them. We found no prior work that bounds how far this metric can move a
control's verdict.

\section{What the anomaly score measures}
\label{sec:score}
\subsection{Data, detector and score}
The data are CERT releases r4.2 and r6.2, which simulate an organization's activity logs and script insider incidents
\citep{certdataset,glasser2013bridging}. An \emph{incident day} is a day in an insider's incident calendar, after our
correction of the released calendars; the label does not imply that every action on the day is malicious. Each user-day is written as text whose tokens
the model must predict. They are of two kinds: \emph{profile targets} (the \texttt{DAY} and \texttt{PSY} lines:
calendar, organizational and simulated personality fields) and \emph{behavior targets} (session counts and activity
lines). The detector is a language model fine-tuned with QLoRA, a parameter-efficient method
\citep{hu2022lora,dettmers2023qlora}, on texts from benign training users; the \emph{base} model is the same model
before fine-tuning. Let $x=(x_1,\ldots,x_T)$ be the token sequence for one user-day. The detector's anomaly score is
its mean next-token surprisal, or loss, over the scored positions $\mathcal T(x)$,
\begin{equation}
 s(x)=\frac{1}{|\mathcal T(x)|}\sum_{t\in\mathcal T(x)}-\log p_\theta(x_t\mid x_{<t}),
 \label{eq:score}
\end{equation}
in nats. Higher values mean that the model predicts the observed text less well; they do not, by definition, mean
malicious intent. Appendix~\ref{app:audit} gives the data, the fine-tuned adapter and the full detail of
Sections~\ref{sec:score}--\ref{sec:interventions}.

\subsection{A score can rank users instead of incident days}
\begin{table}[h!]
\centering\small
\begin{tabular}{lrr}
\toprule
Target losses included & Pooled day AUROC & Within-user AUROC \\
\midrule
All & 0.663 & 0.684 \\
Profile & 0.464 & 0.523 \\
Behavior & 0.843 & 0.851 \\
\bottomrule
\end{tabular}
\caption{Scoring the same predictions three ways on the corrected r4.2 data: pooled AUROC and within-user AUROC, the
latter averaged equally over the 70 insiders.}
\label{tab:scores-main}
\end{table}
Splitting the scored positions into profile and behavior targets decomposes $s$ exactly: it is the
token-count-weighted mean of the two parts' losses. The usual evaluation, pooled AUROC, asks how often an incident day
outscores a normal day of any user. Within-user AUROC instead compares each insider's incident days only with that
insider's own normal days, which is the question an analyst faces when reviewing one person's alerts. On the corrected
r4.2 data, scoring behavior targets alone raises the within-user AUROC, averaged with equal weight per user, from
0.684 to 0.851 (Table~\ref{tab:scores-main}), a difference of $\WithinGain$ with a descriptive 95\% interval from resampling users ($\WithinCI$). It
improves 58 of 70 users. This is better ranking within users on these data, not better detection at an operating
alert threshold. Pooled AUROC hides the difference: 99.53\% of all incident--normal pairs in this data pair days of
different users, so it places little weight on identifying which day of a given insider is an incident. And behavior predictions
still depend on the supplied profile: giving a day (the \emph{receiver} of the edit) another user's profile raises the
fine-tuned model's loss on identical behavior targets, but not the base model's (Appendix~\ref{app:swaps}). A
better-ranked score still does not say what inside the model produced it; Section~\ref{sec:sae} turns to the
features offered as that explanation.

\section{How the explanation's features are chosen}
\label{sec:sae}
\label{sec:ranking}
\subsection{Representing what fine-tuning changed}
An explanation names internal features of the detector, so it depends on how those features are found. At token
position $t$ of the audited layer, let $\delta_t=h_t^{\rm ada}-h_t^{\rm base}$ be the difference between the fine-tuned
and base models' hidden states on the same input, that is, what fine-tuning changed, standardized per coordinate as
$q_t=(\delta_t-\mu)\oslash\sigma$. A TopK sparse autoencoder approximates $q_t$ as a sum of features from a learned
dictionary, keeping only the $k$ largest feature activations $z_t$ at each token \citep{cunningham2023sparse,gao2024scaling}.
Appendix~\ref{app:r62} gives the dictionary's layer, width and sparsity. The explanation is a selected set $S$ of five features, ranked by how much more they activate on the
incident tokens of the \emph{discovery users}, the insiders used for selection, than on a reference population without
insiders.

\subsection{The comparison population changes the explanation}
\begin{figure}[b]
\centering
\includegraphics[width=\linewidth]{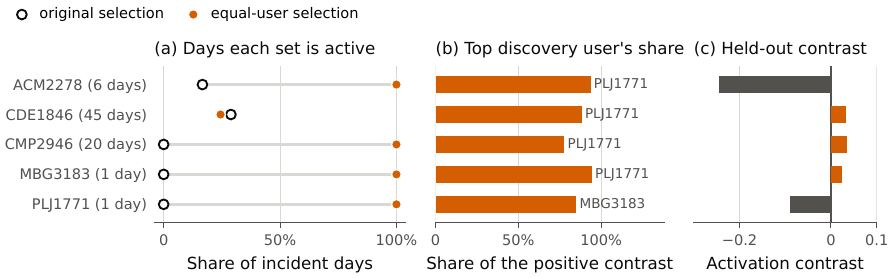}
\caption{The exploratory r6.2 selection by held-out insider, each fold selecting on the other four. (a) The share of the
user's held-out incident days on which the originally selected features and those chosen with equal user weights are
active. (b) The discovery user that supplies most of the equal-user selection's positive contrast, and its share.
(c) That selection's held-out incident-minus-normal activation contrast, which has opposite signs across users.}
\label{fig:selection}
\end{figure}
Such a ranking can select differences between people rather than differences between a person's normal and incident
days. In an exploratory r6.2 analysis that compares each user with themselves, changing the reference population or
giving each user equal weight replaces much of the selection, while weighting by tokens rather than days changes
little. The five features originally selected activate on 14 of 73 held-out incident days, from two
users; the features
chosen with equal user weights activate more widely, but a single discovery user supplies most of their
incident-versus-reference difference, and their contrast on held-out days has opposite signs across users (Figure~\ref{fig:selection}). On r4.2, one
selected feature gets most of its activation among discovery users from a single user. We keep the selection that was
frozen before outcomes and disclose this concentration rather than reselecting after seeing them. That features are
more active on incident days does not show that the model uses them; Section~\ref{sec:interventions} removes them and
measures what changes.

\section{What interventions on the features show}
\label{sec:interventions}
\label{sec:controls}
\subsection{Edits and their controls}
An intervention edits the hidden state of one day, its \emph{receiver}, through the dictionary. Editing a feature set $F$ replaces its activations by
$z'_{t,F}$ and adds the decoded difference to the fine-tuned model's hidden state, keeping the part the SAE does not
reconstruct:
\begin{equation}
 e_t=\sigma\odot D_{:,F}(z'_{t,F}-z_{t,F}),\qquad h'_t=h_t^{\rm ada}+e_t.
 \label{eq:edit}
\end{equation}
A \emph{removal} sets $z'_{t,F}=0$. A \emph{donor patch} moves the receiver day's activations of $F$, where $F$ is
active, toward their average on a donor day, a day of another user; at full dose it moves them all the way. Each edit is compared with
controls, edits that change the state without carrying the explanation: a natural active control $C$, five other features that are often active but barely more so on incident
tokens than on the reference (an activity screen, not equality of edit sizes or edited positions), and random dictionary directions matched in norm to the edit. Effects are in millinats per target
token (1\,mn is $10^{-3}$ nats) and are estimated on 35 confirmation users held out from feature selection,
averaging within each user first and then equally over users; Appendix~\ref{app:donors} gives
the full definitions.

\subsection{Removal shows predictive dependence, not incident specificity}
The experiment covers all 1,362 incident days and 380 normal days matched to them. Removing $S$ raises behavior loss
on incident days by \AblRemBase\,mn (95\% interval \AblRemBaseCI) relative to the unedited model, and by
\AblRemActive\,mn relative to the active control. Measured against that control, however, the removal's effect on
an incident day minus its effect on the matched normal day is \AblIncNorm\,mn \AblIncNormCI: the model relies on
$S$ to predict behavior, but not more so on incident days (Figure~\ref{fig:interventions}a).

\subsection{Benign-donor patches do not show repair}
An earlier pipeline's best-donor result was read as benign users' values of $S$ ``repairing'' an insider's
incident-day behavior predictions. The corrected grid contains all 1,362 incident receiver-days; Appendix~\ref{app:donors} gives its size and the
independent checks of its aggregation. The primary
measure, $\tau$, takes for each receiver the behavior-loss change from patching $S$ with incident donors minus that
from patching it with benign donors, and subtracts the same difference for the control features $C$; the repair
reading predicts $\tau>0$. Instead, $\tau$ is negative in all four organizational contexts used to match donors to
receivers, in both the discovery and the confirmation users, from \DonorTauLo{} to \DonorTauHi\,mn in the latter (Figure~\ref{fig:interventions}b):
full-dose patches of $S$ from benign donors raise behavior loss more than patches from incident donors. The earlier positive best-donor finding is not recovered. This permits a descriptive effect
of the joint set, not a claim that the features encode an insider concept or that benign patches repair behavior.

\subsection{Control fidelity must be checked}
An offline fidelity audit re-encodes the edited hidden states to check what each edit changed. Removal eliminates
every originally active entry of $S$, yet previously inactive members of $S$ activate on some tokens; the random-direction
control keeps most of $S$'s activation mass, pooled over tokens, but amplifies it at many tokens; and the active
control mostly preserves $S$ while editing different positions. These are different control policies, and neither establishes that all nuisance effects are matched
(Appendix~\ref{app:audit}). Interpreting a removal requires a control, and choosing the control requires a metric;
Section~\ref{sec:geometry} measures how much the conclusion depends on that metric.

\section{When the control's metric decides the conclusion}
\label{sec:geometry}
A choice remains even for a
control that preserves the selected features exactly: the metric that decides which edit is closest to the removal
(Section~\ref{sec:geo-choice}). We measure its weight with the tipping radius (Section~\ref{sec:geo-radius}) and test
that radius in each run (Section~\ref{sec:geo-runs}).

\subsection{The overlooked choice: how to measure an edit's size}
\label{sec:geo-choice}
\begin{figure}[t]
\centering
\includegraphics[width=\linewidth]{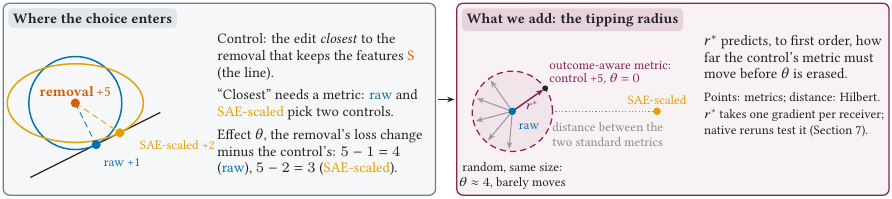}
\caption{The control's metric is the auditor's choice, and the explanation's verdict depends on it (left; illustrative
loss changes, not measurements). The tipping radius $r^\ast$ predicts, to first order, how far that metric must move
before the effect $\theta$ is erased (right); Figure~\ref{fig:runs} tests it in every run.}
\label{fig:overview}
\end{figure}
Consider one removal and two reasonable ways of measuring similarity: one counts every neuron's change equally,
the other relative to its typical size. Each picks its own control, so the same removal yields two estimates of the
features' effect (Figure~\ref{fig:overview}, left). The question is how far the metric must move before that effect
disappears.

We study the choice with a preserving control (Appendix~\ref{app:geometry}).

\begin{definition}[Preserving control and contrast]\label{def:control}
At each targeted token, the \emph{preserving control} under a metric is the edit closest to the realized removal among
the \emph{allowed} edits: those that keep the selected features' encoder readings at their unedited values, give the
other active features their values under the removal, and are no larger than the removal. The \emph{contrast}
$\theta$ is the change in behavior loss under the removal minus that under the control, averaged over 24 small
perturbations of each control.
\end{definition}

``Closest'' requires a metric, as it does for counterfactual explanations \citep{wachter2018counterfactual}. Two were
declared before any outcome was computed: Euclidean distance on the hidden state (raw) and distance weighted by the
SAE's coordinate scales (SAE-scaled). They are the \emph{standard metrics} of the figures and tables. A positive
$\theta$ means that the removal raises the loss more than the control does; a $\theta$ near zero means that the two
edits have the same average effect. Neither sign alone shows what the features encode.

\subsection{The tipping radius: how far the metric must move}
\label{sec:geo-radius}
We consider positive diagonal metrics, which weight each coordinate separately, and measure the distance between two
of them by the Hilbert projective distance $d_H$. Its exponential, $e^{d_H}$, is the largest factor by which the two metrics disagree
about the relative cost of two coordinates, and hence of any two edits, since a metric change rescales an edit's cost
by a weighted mean of its coordinates' factors. We report distances as a share of the \emph{gap}, the distance between the two
standard metrics, so that 10\% is a tenth of the way from one to the other.

Let $c$, the \emph{rate vector}, collect the first-order derivatives of $\theta$ with respect to the metric's
log-weights, obtained from one gradient per receiver and the control's construction. Within distance $r$ of the
declared metric, the metric that is worst for $\theta$, the \emph{outcome-aware} metric (finding it requires the
outcome), changes it by $(r/2)\|c\|_1$ to first order.

\begin{definition}[Tipping radius]\label{def:rstar}
The \emph{tipping radius} is the Hilbert distance from the declared metric at which the outcome-aware metric erases
the contrast to first order:
\begin{equation}
 r^\ast=\frac{2|\theta|}{\|c\|_1},
 \label{eq:rstar}
\end{equation}
with $r^\ast=+\infty$ if $c=0$, when no metric change moves $\theta$ to first order.
\end{definition}

Figure~\ref{fig:overview} (right) sketches it. Like an E-value \citep{vanderweele2017evalue}, $r^\ast$ states how
strong an unfavorable choice would have to be before the conclusion changes. It is a first-order quantity, a prediction
rather than a certificate, so we test it natively (Section~\ref{sec:geo-runs}). For a population we apply the same
construction to the equal-user mean contrast and a single metric shared by all receivers.

A \emph{random metric}, moved the same distance with random signs, moves $\theta$ by an amount of order
$(r/2)\|c\|_2$ instead. At $r^\ast$ that is about $|\theta|/\sqrt{d_{\rm eff}}$, where
\begin{equation}
 d_{\rm eff}=\left(\frac{\|c\|_1}{\|c\|_2}\right)^{2}\le d
 \label{eq:deff}
\end{equation}
is the effective number of coordinates of $c$ (Proposition~\ref{prop:gap}). When the rate vector is spread over many
coordinates, $d_{\rm eff}$ is large and random metrics miss the direction that erases $\theta$; when it is
concentrated on a few, random checks can already reveal the sensitivity. Random metrics stay uninformative up to
$G=\sqrt{d_{\rm eff}}/z$ times $r^\ast$, with $z=\sqrt{2\ln40}$. Where the declared rules test it, they require
$G\ge5$, for the population or the median receiver (the \emph{spread requirement}); a \emph{site}, the layer whose
coordinates the metric weights, with 64 coordinates has $G\le2.9$. On the detectors, $c$ is spread over many
coordinates. There, random metrics erase $\theta$ with probability at most 0.05 up to a distance that is
\CtlOutRatioMin--\CtlOutRatioMax{} times $r^\ast$ (medians over receivers: all of the 8B's, and the 3B's whose
contrasts exceed three standard errors).

\subsection{How each run tests the radius, and how we read it}
\label{sec:geo-runs}
The runs answer three questions. Does the tipping radius predict what actually happens? Each run rebuilds the controls
under the outcome-aware metric and reruns the model. Would an ordinary robustness check find the same weakness? Each run
compares that metric with random metrics at the same distance. Does the result generalize? Runs repeat the test on
other models, data and fresh documents, test whether a metric fixed on some documents transfers to others, and retrain
detectors.

Each run declares its two metrics, its population, the tipping radius's first-order predictions and a verdict rule
before its outcomes are computed. It then measures the population contrast natively under the outcome-aware metric at
$r^\ast$, under two random metrics at the same distance, and under the other standard metric (definitions and tests in
Appendix~\ref{app:geometry}). \emph{First order} means predicted from one gradient per receiver, a linear
approximation; \emph{native} results rebuild the controls under the changed metric and rerun the model. A result read
by a rule fixed before the run is marked \emph{declared}, and the others post hoc or exploratory; a run's verdict is
its declared rule's reading (Appendices~\ref{app:geometry}--\ref{app:xseven}).

Two of the declared tests recur in Section~\ref{sec:results}. The \emph{population test} applies the construction to
the effect averaged over users, with one metric shared by all their receivers, and checks that the outcome-aware and
random metrics change it as first order predicts. The \emph{radius test} checks individual receivers: among those whose
effects reach 1\,mn, of which at least 10 are needed, the median $r^\ast$ must be at most 0.5 (a factor of
$e^{0.5}\approx1.65$ in relative cost), and three further predictions must hold.

\paragraph{Settings and runs.}
We ran the test on \CtlSetRuns{} runs of \CtlSetModels{} language models from five families (counting DeepSeek's
distillation of Qwen as Qwen), in four settings (Table~\ref{tab:settings}; Figure~\ref{fig:runs}): language-model
detectors adapted to CERT activity logs, and openly released language models with openly released SAEs (\emph{public
models}) reading CERT days, web text, and the TWOS and LANL security logs. The security logs are short stretches of activity from TWOS, whose 24
participants competed under incentives for both normal and malicious behavior, including masquerade and traitor attacks
\citep{harilal2017twos}, and windows of LANL's de-identified 2015 authentication events
\citep{kent2016cybersecurity,kent2015comprehensive}, serialized as text and read as released. Each run is read at both standard metrics, giving \CtlSetPoints{} run--metric pairs. The
exploratory 8B and Pythia's float64 rerun are not declared runs; each replication on fresh documents is one.
Appendix~\ref{app:runs} lists the runs, including four declared runs that contribute no pairs. For
the public models each receiver is a document: a web-text passage, a CERT day or a log window. The detectors that are
not language models are classic autoencoder anomaly detectors, unrelated to the sparse autoencoders that explain the
language models, on CERT days and on the TWOS and LANL windows (Appendix~\ref{app:neural}).

\paragraph{Reading each run.}
Each run gives three numbers at each of its two standard metrics (a run--metric pair), all relative to its declared
contrast $\theta$: the distance $r^\ast$ of the outcome-aware metric, as a share of the gap; the \emph{residual}, the
share of $\theta$ left under that metric (100\% is unchanged, 0 erased and below 0 reversed); and the larger change that
two random metrics at the same distance make to $\theta$, the check an auditor can run without knowing the outcome.
Uncertainty uses the setting's sampling unit: users for the detectors, documents for web text, and documents clustered
by user for the security logs and the CERT days. A pair counts as \emph{erased} if the one-sided 95\% upper bound on
the share that remains is at most 20\%, the threshold declared for the web-text replications and applied post hoc
elsewhere, and as \emph{reversed} if that bound is below zero. Table~\ref{tab:settings} gives, for each setting and each detector, the
median and range of the first two numbers over its pairs, how many pairs were erased, the largest random change, and
how many of its declared runs' readings held; Table~\ref{tab:runs} in Appendix~\ref{app:runs} gives every run.

\begin{table}[t]
\centering
{\footnotesize\setlength{\tabcolsep}{4pt}\begin{tabular}{@{}lrrllrrr@{}}
\toprule
Setting & Runs & Pairs & $r^\ast$/gap (\%) & Residual (\% of $\theta$) & Erased & Random (\%) & Declared, held \\
\midrule
\multicolumn{8}{@{}l}{\textit{Detectors fine-tuned on CERT}} \\
\quad Qwen3-8B (exploratory) & 1 & 2 & 1.9 [1.8, 2.0] & $-$36 [$-$37, $-$36] & 2 of 2 & 2.8 & -- \\
\quad Qwen2.5-3B, cohort 1 & 1 & 2 & 5.8 [4.9, 6.6] & $-$28 [$-$37, $-$18] & 0 of 2 & 8.8 & 0 of 1 \\
\quad Qwen2.5-3B, cohort 2 & 1 & 2 & 5.4 [4.8, 5.9] & 17 [3, 32] & 0 of 2 & 7.4 & 1 of 1 \\
\quad Qwen2.5-3B, seeds 43--45 & 3 & 6 & 2.8 [2.1, 15.0] & 29 [$-$6, 42] & 0 of 6 & 17.4 & 1 of 3 \\
\quad Gemma-3-1B & 1 & 2 & 83.6 [74.0, 93.1] & 68 [39, 96] & 0 of 2 & 18.6 & 0 of 1 \\
\quad Llama-3.2-1B & 1 & -- & \multicolumn{4}{l}{3 of 600 admitted, too few to test} & 0 of 1 \\
Public models, CERT days & 3 & 6 & 4.3 [2.2, 5.3] & 0 [$-$3, 0] & 6 of 6 & 6.9 & 3 of 3 \\
Public models, web text & 6 & 10 & 11.7 [2.0, 17.6] & $-$7 [$-$65, 5] & 8 of 10 & 10.0 & 4 of 5 \\
Public models, security logs & 18 & 36 & 5.3 [1.6, 6.7] & 0 [$-$1, 2] & 36 of 36 & 11.0 & 16 of 18 \\
\multicolumn{8}{@{}l}{\textit{Detectors fine-tuned on TWOS and LANL}} \\
\quad Qwen2.5-3B, TWOS, seeds 42--43 & 2 & -- & \multicolumn{4}{l}{radius test only (12 users, under 25)} & 2 of 2 \\
\quad Qwen2.5-3B, LANL & 1 & 2 & 8.8 [7.8, 9.7] & $-$9 [$-$9, $-$8] & 0 of 2 & 27.5 & 0 of 1 \\
Autoencoders (CERT, TWOS, LANL) & 6 & 4 & 34.7 [16.6, 73.0] & $-$3 [$-$35, 19] & 3 of 4 & 34.7 & 0 of 6 \\
\bottomrule
\end{tabular}}
\caption{The test in each setting, one row per detector: medians [ranges] over run--metric pairs (point estimates). Runs counts every run,
including four declared runs that yield no pairs (Section~\ref{sec:geo-runs}).}
\label{tab:settings}
\end{table}

\section{Results}
\label{sec:results}
Table~\ref{tab:settings} summarizes each setting and detector, Figure~\ref{fig:runs} each run; the later figures test the radius.

\subsection{A change smaller than the gap between the standard metrics removes most of the contrast}
\label{sec:res-effect}
\begin{figure}[t]
\centering
\includegraphics[width=\linewidth]{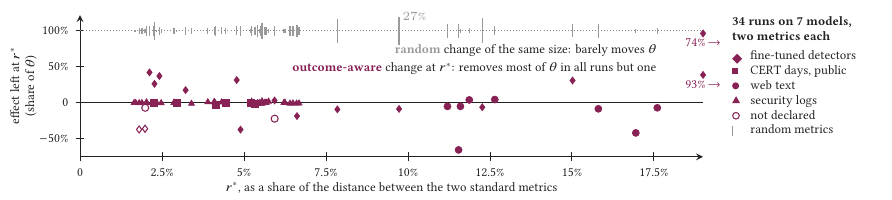}
\caption{Every run: the effect left under the outcome-aware metric at $r^\ast$, against $r^\ast$ as a share of the gap.
Grey bars: the larger change from two random metrics at that distance, cut at $\pm$20\% with the value. Points beyond
the axis sit at its edge with their share (Table~\ref{tab:runs}).}
\label{fig:runs}
\end{figure}

In \CtlSetRunsCore{} of the \CtlSetRuns{} language-model runs, the outcome-aware metric lay at most \CtlSetGapMax\% of
the gap from the declared metric and removed at least \CtlSetRemovedMin\% of the contrast (Figure~\ref{fig:runs}).
It changed the relative cost of any two edits by a factor of at most \CtlSetCostMin{} to \CtlSetCostMax, depending on
the run. With sampling error, the contrast was erased in \CtlResHold{} of the \CtlSetPoints{} pairs: every security-log
and CERT-day pair, \CtlResWebHold{} of the \CtlResWebPairs{} web-text pairs, and the exploratory 8B's two. With their
users as the sample, the other detector pairs were not resolved: their intervals were too wide to show erasure. The
exception was the Gemma-3-1B detector, whose small, negative contrast needed metrics \CtlXSixGemmaGapRaw\% and
\CtlXSixGemmaGapSae\% of the gap away, which removed \CtlXSixGemmaRemovedRaw\% and \CtlXSixGemmaRemovedSae\% of it (Appendix~\ref{app:xsix}).

By the rules declared before each run, the effect replicated in \CtlPubDeclYes{} of the \CtlPubDeclN{} primary public-model runs;
the other, Pythia's float32 run on web text, stopped at its first-order gate for a numerical reason
(Appendix~\ref{app:public}). The 3B detector was tested on two declared samples of receivers (cohorts 1 and 2). The
population predictions held for both, but cohort 1's declared verdict was \emph{inconsistent}, because most individual
receivers were far from fragile. Cohort 2's radius test, of a claim formed after cohort 1's result, was supported
(Appendix~\ref{app:geometry}); Section~\ref{sec:res-seeds} tests this radius result on other detectors.

\textbf{Observation 1.} The test can report robustness as well as fragility, but in \CtlSetRunsCore{} of the
\CtlSetRuns{} runs an effect attributed to the explained features turned on a change of metric at most
\CtlSetGapMax\% of the gap.

\subsection{Ablation: without the outcome, a change of the same size leaves the effect intact}
\label{sec:res-blind}
\begin{figure}[t]
\centering
\includegraphics[width=\linewidth]{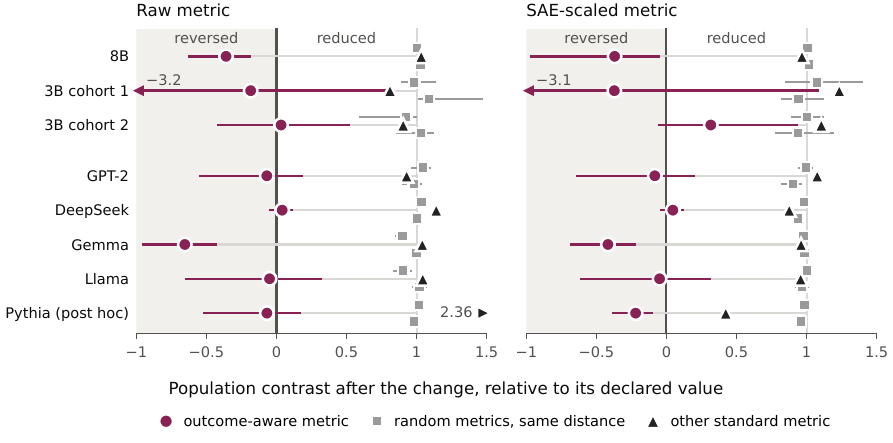}
\caption{A same-size outcome-aware change removes most of the population contrast or reverses it, while random
metrics at that distance and the other standard metric keep its sign. One row per run, at each standard metric
(panels). Horizontal axis: the contrast after the
change, relative to its declared value (1 unchanged, 0 erased, below 0 reversed); arrowheads mark values beyond the
axis. Bars: 95\% intervals for these ratios (Fieller's method); the 3B's are wide because its contrasts lie only \CtlResThreeThetaTMin--\CtlResThreeThetaTMax{} standard errors
from zero.}
\label{fig:false-assurance}
\end{figure}

To isolate what the outcome-aware direction contributes, we kept the size of the change at $r^\ast$ and replaced its
direction with ones chosen without the outcome (Figure~\ref{fig:false-assurance}). Random metrics at that distance moved
the contrast by at most \CtlSetBlindMaxNoSeed\%, each keeping at least \CtlResRandLow\% of it with sampling error
(one-sided 95\%, post hoc); a web-text replication block, outside these runs, reached \CtlWebRepRandMax\%
(Appendix~\ref{app:public}). To first order, none of \CtlAblDraws{} random metrics per pair erased it in any of the
\CtlAblPairs{} pairs of Table~\ref{tab:sensitivity}. Moving the same distance toward the other standard metric changed
the contrast by \CtlAblLmPathMin\% to \CtlAblLmPathMax\%. Switching to that metric outright kept every contrast's sign
and changed its size by \CtlOtherAllMin--\CtlOtherAllMax\% in the 8B, the 3B's two cohorts, the declared public-model
runs and the security-log replications; in Pythia's post hoc rerun the two metrics' contrasts differ by a factor of
\CtlPyOtherFactor{}, not in sign (Appendix~\ref{app:public}). The erasure therefore comes from the direction of the
change, not from its size.

Five detector runs are exceptions, all with small or unresolved contrasts. In the 3B's three seeds, at one metric of each
the contrast is not distinguishable from zero across its users; the LANL detector's contrast is not distinguishable
from zero at either metric; and the Gemma-3-1B detector's contrast is small. Random metrics moved these contrasts by
up to \CtlSetBlindMax\%, within the declared bounds, but their random bounds (the one-sided lower bounds on how much of
$\theta$ a random metric keeps) are weak or undefined (Appendix~\ref{app:geometry}).

\textbf{Observation 2.} The random checks we tested, which an auditor could run without the outcome, would have
reported these effects as robust.

\subsection{First order tracks the native response, but $r^\ast$ is not a guarantee}
\label{sec:res-native}
\begin{figure}[t]
\centering
\includegraphics[width=\linewidth]{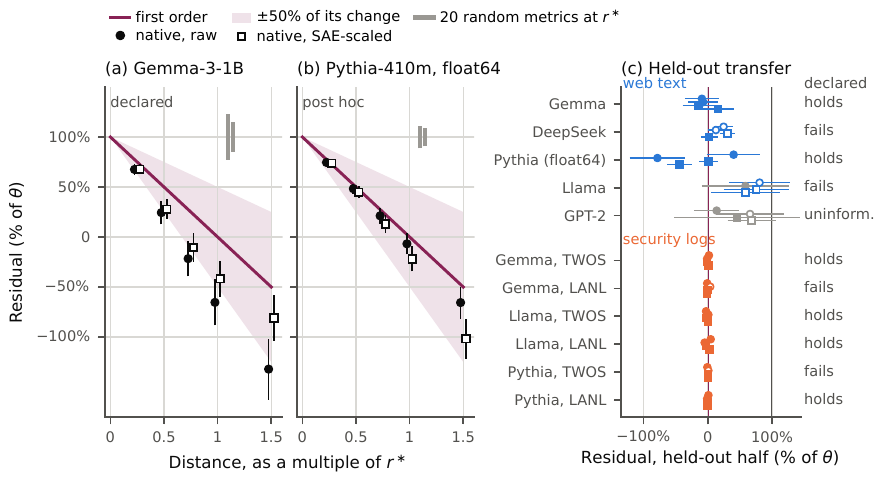}
\caption{The first-order radius holds up natively but understates the response, and a fixed metric transfers only
partly. (a, b) Declared response curves: the residual as the metric moves toward the outcome-aware one; line: first
order; band: $\pm$50\% of its predicted change (the declared tolerance adds two standard errors); points: native,
$\pm$2 SE; grey: the range of 20 random metrics at
$r^\ast$. (c) Declared split-sample test: a metric built on half of the documents, applied to the other half in both
directions at both metrics, $\pm$2 SE; filled: the erasure prediction held; open: it
failed; grey: not informative. Right: each run's declared verdict.}
\label{fig:validation}
\end{figure}

Rebuilding the controls and rerunning the model shows that first order predicts the native change well: at the
population radius the native change was \CtlRatioMin--\CtlRatioMax{} times its first-order prediction on the exploratory
8B and the 3B's cohorts and \CtlSecRatioMin--\CtlSecRatioMax{} times on the security logs, and the declared per-receiver
check of first order passed in \CtlValidatedTests{} of six model--metric tests (Appendix~\ref{app:geometry}). Along the
declared response curves of Gemma and Pythia's float64 run (Figure~\ref{fig:validation}a--b), every native change stayed
within the declared tolerance of first order but ran larger, so the contrast crossed zero before the radius (declared
verdicts: Gemma \emph{\CtlPMFiveGmVerdict}, Pythia \emph{\CtlPMFivePyVerdict}). First order thus erred toward overstating
robustness, as it did in all detector pairs but four. But $r^\ast$ is not a guarantee: at least 0.5\% of the contrast
remained at $r^\ast$ in \CtlResLeftPairs{} of the \CtlSetPoints{} pairs (Table~\ref{tab:runs}), among them DeepSeek's,
whose native changes fell slightly short of first order.

\textbf{Observation 3.} The radius usually overstates how far the metric must move, but it is not a certificate.

\subsection{Rerunning the test on new documents differs from transferring a fixed metric}
\label{sec:res-replication}
\label{sec:res-transfer}
Rerunning recomputes the test on fresh documents; transfer applies a metric fixed on one set of documents to another.

On the security logs, \CtlSecRepYes{} of the \CtlSecRepN{} replications on fresh documents replicated. The two that did
not replicate left \CtlSecRepLeftMin--\CtlSecRepLeftMax\% of the contrast at one metric, a remainder detectable only
because those contrasts lie \CtlSecRepLeftThetaZMin--\CtlSecRepLeftThetaZMax{} standard errors from zero between
documents (Appendix~\ref{app:logs}). On two further blocks of web text, Gemma's, Llama's and Pythia's (float64) runs replicated in
every block, and their stricter declared reading held for Llama and Pythia but not for Gemma (Appendix~\ref{app:public}).

A metric fixed on half of the documents transferred to the other half on the security logs, leaving
\CtlSecXsResidMin\% to \CtlSecXsResidMax\% of the contrast in every direction. It transferred only partly on web text,
where it left up to \CtlXsLlResidMax\% in Llama; Figure~\ref{fig:validation}c gives each run's declared verdict.

\textbf{Observation 4.} Rerunning the test on fresh documents mostly finds the effect again; a metric fixed on one half
of the documents transfers to the other on the security logs, but only partly on web text.

\subsection{Training seeds and detector families give mixed confirmation}
\label{sec:res-seeds}
Retraining asks whether the result holds for a different learned detector, not only for different documents. Apart
from the 3B, each detector is a single training run. Of \CtlXFiveSeedsN{} further 3B seeds, \CtlXFiveSeedsRep{}
(seed~44) replicated seed~42's result. Seeds 43 and 45 were uninformative rather than contrary, with only
\CtlXFiveUninfMin--\CtlXFiveUninfMax{} receivers at 1\,mn against the 10 required (Figure~\ref{fig:coverage}).
Llama-3.2-1B and Gemma-3-1B detectors trained by the same recipe did not replicate it: the Llama detector admitted (could build controls for) only
\CtlXSixLlamaAdmitted{} of the \CtlXFiveOf{} receivers, too few for either test, and the Gemma detector's radius test
was \CtlXSixGemmaXFourVerdict{} (Appendix~\ref{app:xsix}).

\paragraph{Detectors fine-tuned on real activity logs.}
\label{sec:res-twos}
Two 3B adapters fine-tuned on TWOS (seeds 42 and 43) were tested on 600 training windows, 50 from each of TWOS's 12
training users. Both radius tests were \CtlXSevenTwosAXFourVerdict, with median per-receiver $r^\ast$
\CtlXSevenTwosAHOneMedRaw{} and \CtlXSevenTwosAHOneMedSae{} (seed 42) and \CtlXSevenTwosBHOneMedRaw{} and
\CtlXSevenTwosBHOneMedSae{} (seed 43) against 0.5, so both seeds replicated the 3B's radius result; unlike on CERT, small
contrasts were fragile too. With 12 users the population test could not be informative (they need 25 users). The 3B adapter fine-tuned on LANL's authentication logs did not replicate it: only one receiver per metric reached
1\,mn, too few for the radius test, and its population contrast, though reversed by the outcome-aware metric, was not
distinguishable from zero across users (Appendix~\ref{app:xseven}).

\textbf{Observation 5.} Where a detector did not replicate the radius result, it lacked receivers with measurable
contrasts rather than contradicting it.

\subsection{Sensitivity to the method's own choices}
\label{sec:sensitivity}
We varied three of the test's own choices after the fact, on the committed records of \CtlAblRuns{} runs
(Table~\ref{tab:sensitivity} in Appendix~\ref{app:geometry}); these variations do not enter the verdict counts above.

\begin{figure}[t]
\centering
\includegraphics[width=\linewidth]{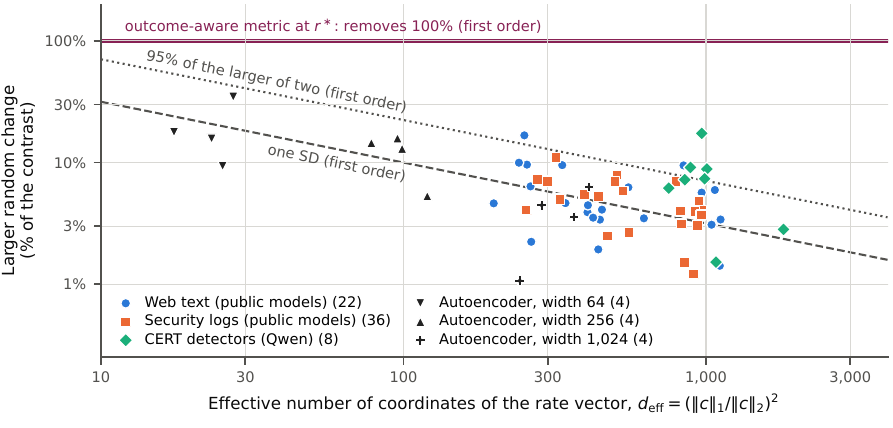}
\caption{Random checks are blind only when the rate vector is spread out. Each point is a run--metric pair: the larger
of its two random changes at $r^\ast$ against $d_{\mathrm{eff}}$. Lines: first order's SD, $1/\sqrt{d_{\mathrm{eff}}}$, and
95\% bound, $\CtlAblQMaxTwo/\sqrt{d_{\mathrm{eff}}}$. The autoencoder at widths 256 and 1,024 is the declared width sweep. Post hoc.}
\label{fig:sensitivity}
\end{figure}

\paragraph{The scale of ``small''.}
In $\ell_2$ on the log-weights (the Euclidean distance between the two metrics' log-weight vectors) instead of the
Hilbert scale, the erasing metric lay \CtlAblLmLtwoMin--\CtlAblLmLtwoMax\% of the way between the two standard metrics
in the language-model pairs of Table~\ref{tab:sensitivity}. In \CtlAblLtwoOver{} web-text pairs it lay at least as far
from the declared metric as the other standard metric. The radius is small in one precise sense: it changes no
relative edit cost by more than a factor $e^{r^\ast}$.

\paragraph{The spread of the rate vector.}
The blind spot of Section~\ref{sec:res-blind} tracked how far the rate vector was spread (Figure~\ref{fig:sensitivity}).
Across the language-model pairs, $d_{\mathrm{eff}}$ was \CtlAblLmDeffMin--\CtlAblLmDeffMax, so a random change at
$r^\ast$ had a first-order standard deviation of at most \CtlAblLmSdMax\% of $\theta$, and first order described the
measured random changes (Appendix~\ref{app:geometry}). An autoencoder detector, with
\CtlAblAeDeffMin--\CtlAblAeDeffMax{} effective coordinates at width 64, is the counterexample. Its outcome-aware metric
still removed most of the contrast or reversed it, but random metrics at the same distance moved it by up to
\CtlAeBlindMax\%, which an auditor would notice, and its site is too small for the spread requirement
(Section~\ref{sec:geo-radius}).

\textbf{Observation 6.} Random checks are informative only where the effect is concentrated on few coordinates. These
results cover diagonal metrics and the receivers whose controls could be constructed.

\section{What an alert's explanation can and cannot support}
\label{sec:implications}
Table~\ref{tab:inferences} links each finding to the inference it licenses for an audit and to the stronger
inference it does not. The rows are conclusions from our measurements, not a validated decision aid: we studied
no analysts, organizations or workers.

\begin{table}[tbp]
\centering\footnotesize
\begin{tabularx}{\linewidth}{@{}>{\raggedright\arraybackslash}p{3.6cm}>{\raggedright\arraybackslash}X>{\raggedright\arraybackslash}X@{}}
\toprule
Finding & Defensible implication for an audit & What it does not establish \\
\midrule
Behavior-target scoring improves within-user ranking (Section~\ref{sec:score}) & Evaluate the task the alert is
meant to represent; report within-user incident discrimination separately from pooled ranking & Fewer
investigations, fewer false accusations, or better outcomes at an operational threshold \\
Behavior predictions still depend on profile inputs (Section~\ref{sec:score}) & Disclose which inputs remain
available after changing the score & That excluding profile losses removes profile information, or that all role
dependence is illegitimate \\
Feature rankings depend on reference users and weighting (Section~\ref{sec:ranking}) & Record the comparison
population and concentration; do not present one user's feature as a general insider mechanism & A demonstrated
demographic fairness violation \\
Donor and control choices change intervention interpretations (Section~\ref{sec:interventions}) & Preserve the estimand, audit control fidelity, and disclose selection and exclusion rules &
A universal explanation, a benign ``repair'', or a uniquely correct control \\
A small outcome-aware change in the control's metric erases or reverses an effect averaged over users; random
changes of the same size barely move it (Sections~\ref{sec:geometry}--\ref{sec:results}) & Fix and disclose the control's metric before outcomes are computed;
report $r^\ast$ and the outcome-blind spread with each effect & That the declared metric is the right one, or a
certified bound on the native effect \\
Controls can be built only for some receivers (Section~\ref{sec:results}) & Report whose cases receive an
explanation and which are excluded & A valid explanation for every alert, or equal access to explanations across
social groups \\
\bottomrule
\end{tabularx}
\caption{Inferences our measurements support for auditing detector explanations, and those they do not.}
\label{tab:inferences}
\end{table}

Together, these findings suggest that a claim that a detector's internal features explain its alerts should report
the scoring view, the reference population for feature selection, the control policy with its metric and tipping
radius, and the cases for which no explanation could be built.
These are proposed practices, not a tested
intervention: whether they improve decisions must itself be tested, as work on explanations for discrimination
detection shows \citep{skirzynski2025discrimination}.

\section{Limits of the evidence}
\label{sec:related}
Both CERT datasets are synthetic, and r6.2's five users do not support broad population claims. The security-log runs
use real logs but test only the control's metric, with general-purpose language models and with three 3B detectors
fine-tuned on those logs (two on TWOS, one on LANL).
Confirmation users we had inspected earlier served as development data for later method design. Our user-resampling
intervals condition on fixed representations and donors; uncertainty from training, selection and the donor
population is outside their scope, and reused donors matter for broader inference.

\section{Conclusion}
A score, a feature explanation and an intervention answer different questions. Scoring behavior alone ranks an
insider's incident days better within users, though profile information still shapes the model's predictions; the
features named depend on whom they are compared against; and corrected donor patches do not show repair. Even a control
that preserves the named features exactly depends on its metric: a small change to it, chosen with knowledge of the
outcome, can erase or reverse an effect averaged over users, while random changes of the same size usually leave it
intact. Declaring that metric before outcomes and reporting its tipping radius are part of the evidence an audit gives. The radius is a
first-order prediction for one choice among several, not a guarantee. Still, an explanation that does not report how
its test was built cannot tell an analyst, or the worker it concerns, what an alert rests on.

\clearpage
\section*{Ethical Considerations Statement}
This work uses the CERT r4.2 and r6.2 insider-threat datasets. They are synthetic: they describe simulated employees,
and their incidents are scripted. Appendix~\ref{app:logs} also uses two public datasets that record real activity:
TWOS, collected from 24 participants in a competition held to produce it \citep{harilal2017twos}, and LANL's 2015
authentication events, which the laboratory de-identified and released under CC0 \citep{kent2015comprehensive}. We use
both as released, report only aggregate statistics of public language models' behavior on them, and make no attempt to
identify anyone. We collected no data from people. No ethics review was required, and the terms of use of the CERT,
TWOS and LANL data permit this research use with citation.

The principal risk we considered is that an interpretability audit of a workplace-monitoring model could be read as
endorsing such monitoring, or as validating allegations against workers. We address it by separating claims about
prediction from claims about internal mechanisms and from claims about people, by reporting negative results and
retired claims, and by stating what each finding does not establish (Section~\ref{sec:implications}). A second risk
is dual use. The control-geometry analysis shows how a small analysis choice, made with knowledge of the outcome, can
erase or reverse a conclusion, which could guide manipulation as well as auditing. We report it because declaring the
control's metric in advance, together with the tipping radius, makes such manipulation detectable and costly.
The fine-tuned adapters and the SAEs will be released.

\section*{Generative AI Usage Statement}
Under our direction, Anthropic's Claude (including Claude Opus 5.5) and OpenAI's GPT-6 Astra wrote and ran analysis
and experiment code and helped draft and edit the manuscript. Every reported number is rendered from hash-bound evidence files by scripts that verify those hashes, and
the analysis code was tested with deliberate mutations. We directed the work, reviewed the results and the text, and
take full responsibility for the content.

\begin{acks}
This work used Anvil at Purdue University through allocation CIS230270 from the Advanced Cyberinfrastructure Coordination Ecosystem: Services \& Support (ACCESS) program~\cite{boerner2023access}, which is supported by U.S. National Science Foundation grants \#2138259, \#2138286, \#2138307, \#2137603, and \#2138296. It also used the Magnolia cluster at The University of Southern Mississippi, supported by the National Science Foundation's Major Research Instrumentation program under grant ACI-1626217.
\end{acks}

\bibliographystyle{ACM-Reference-Format}
\bibliography{references}

\appendix

\FloatBarrier\section{The detector audit: definitions and evidence}
\label{app:audit}

\subsection{Data, targets and the score}

\paragraph{Data and the meaning of a label.}
CERT releases r4.2 and r6.2 contain simulated organizational activity and
scripted insider incidents \citep{certdataset,glasser2013bridging}.
An \emph{insider} is a user identified by the official answers; an
\emph{incident day} is a day in that user's corrected incident calendar.
This day label does not imply that every action or token on the day is malicious.
A normal day of an insider is not a benign-user identity. The distinction
matters both for training exclusion and for donor eligibility.

Each user-day contains a \texttt{DAY} header with calendar/organizational
fields, a \texttt{PSY} line with simulated personality attributes, a session
count, and \texttt{SES} lines with activity counts. We call
\texttt{DAY}/\texttt{PSY} targets \emph{profile} targets. All other scored
targets---including session-count, session, separator and special-token
classes---form the complementary \emph{behavior} view. These names specify
input locations; not every profile field is static, and neither class name
certifies what an internal feature computes.

The corrected r4.2 scoring cohort contains 42,468 days from 149 users:
70 insiders and 79 non-insider comparison users, with 1,362 incident days.
The r4.2 donor experiment uses those 1,362 incident days, split into 35
discovery users (695 days) and 35 confirmation users (667 days).
The r6.2 incident cohort contains five users and 73 incident days, distributed
as 6, 45, 20, 1 and 1. Each r6.2 insider represents a different scenario;
r4.2 has 30, 30 and 10 insiders in its three scenarios.

Training excludes insider identities. The non-insider comparison users in
the principal r4.2 scoring cohort are held out from adaptation. This is not
the same evaluation as comparing incidents against a mixture of seen and
unseen benign users. All findings below retain their stated cohort.

\paragraph{An exact decomposition, with a limited interpretation.}
Partition $\mathcal T(x)$ into profile positions $P(x)$ and behavior positions
$B(x)$. For $V\in\{P,B\}$, let $n_V=|V(x)|$ and
$s_V(x)=n_V^{-1}\sum_{t\in V(x)}\ell_t(x)$ when $n_V>0$. Then
\begin{equation}
 s(x)=\frac{n_P}{n_P+n_B}s_P(x)+\frac{n_B}{n_P+n_B}s_B(x).
 \label{eq:decomposition}
\end{equation}
This is a token-count-weighted identity for each day, not a fitted explanation.
An empty view is unavailable. Changing the score from $s$ to $s_B$ leaves
the input and every prediction unchanged. In particular, a behavior target
can still depend on the preceding profile. Excluding profile losses from the
score is different from deleting the profile text or retraining without it.

\paragraph{Distinguishing a user from an incident day.}
AUROC counts how often a positive example receives a higher score than a
negative example, assigning half credit to a tie. For positive and negative
day sets $\mathcal I$ and $\mathcal N$, define
\begin{equation}
 H(a,b)=\ind[a>b]+\tfrac12\ind[a=b],\qquad
 {\rm AUC}(s)=\frac{\sum_{i\in\mathcal I}\sum_{j\in\mathcal N}H(s_i,s_j)}
 {|\mathcal I||\mathcal N|}.
\end{equation}
Here $s_i$ is the score of day $i$ and $\ind[\cdot]$ is one when its condition
holds and zero otherwise. Partition these day pairs by whether they belong
to the same user. If $w$ is the fraction of same-user pairs, the identity
\begin{equation}
 {\rm AUC}_{\rm pooled}=w\,{\rm AUC}_{\rm within,pair}
                       +(1-w)\,{\rm AUC}_{\rm cross}
 \label{eq:auc}
\end{equation}
separates pair-weighted within-user discrimination from cross-user placement.
The equal-user mean of individual within-user AUROCs is a different summary;
it gives each eligible insider equal weight.

\label{app:swaps}
\paragraph{Profile-conditioned behavior predictions.}
A separate input-swap experiment tests whether behavior predictions depend
on the supplied profile. For every incident receiver-day, a frozen donor
policy chooses an official non-insider validation user and one of their days.
Static organizational fields and personality attributes are replaced while
the receiver's week and behavior are retained. The comparison scores identical
behavior targets, aligned by character span and token identity, with the
adapter enabled and disabled on the same quantized backbone.

The equal-user profile-swap loss increase is 0.1535 nats/target for the adapted
r4.2 model and $-0.0008$ for its base counterpart. Their difference is
0.1543, with a descriptive user-resampling interval $[0.1338,0.1748]$.
The r6.2 difference is 0.1250 on five users. These results support dependence
of behavior predictions on the supplied profile under this input policy;
they do not establish that such conditioning is harmful or mediated by the
selected SAE features. Changing a profile can also move token positions.

The corrected input-swap experiment covers 1,362 r4.2 and 73 r6.2 incident
days. Each receives one fixed donor, selected by hash by first choosing an
official non-insider validation user and then one of that user's days.
The experiment scores original, profile-swapped and behavior-swapped texts
with the adapter enabled and disabled, for 8,610 rows. No input is truncated
and no aligned behavior target is dropped. A swapped text has no asserted
incident ground truth.

\begin{table}[ht]
\centering\begin{tabular}{lrrrrr}
\toprule
Release & Users & Adapted & Base & Difference & 95\% interval \\
\midrule
r4.2 & 70 & +0.1535 & -0.0008 & +0.1543 & [+0.1338, +0.1748] \\
r6.2 & 5 & +0.1158 & -0.0093 & +0.1250 & [+0.0864, +0.1637] \\
\bottomrule
\end{tabular}
\caption{Profile-swap minus original loss on identical behavior targets,
in nats/target. Means weight receiver users equally. Difference is adapted
minus base, and its interval resamples receiver users with donors fixed.}
\end{table}

Profile replacement changes absolute behavior-token positions for 768 r4.2
and 45 r6.2 days. A post hoc unchanged-position restriction retains 594
days/63 users and 28 days/five users, respectively; the adapted-minus-base
contrasts remain positive (0.1670 and 0.1121). This is a sensitivity on a
different subset, not independent confirmation. Relative importance cannot
be inferred by comparing profile swaps with behavior swaps: the latter also
change the prediction targets, and equal-day versus equal-user weighting
reverses their full-score ordering in these runs.

\subsection{Recorded main-adapter configuration}
The frozen main adapters use \texttt{Qwen/Qwen3-8B}. Their saved training
summaries give Table~\ref{tab:training}: gradient accumulation is one,
the seed is 42, the maximum sequence length is 2,048, and gradient
checkpointing is enabled. Both records specify no evaluation during training;
the listed validation rows do not imply checkpoint selection or early stopping.
The saved r4.2 adapter configuration gives rank 16, scaling $\gamma=32$,
dropout 0.05 and no adapter bias, on the query, key, value, output, gate,
up and down projection matrices. Evaluation masks introduced for the audit
do not change the already-trained adapter.

\begin{table}[ht]
\centering\begin{tabular}{lrrrrr}
\toprule
Release & Training days & Validation days & Devices & Rows/device & Rows/step \\
\midrule
r4.2 & 287,827 & 27,026 & 4 & 22 & 88 \\
r6.2 & 1,251,225 & 140,711 & 4 & 12 & 48 \\
\bottomrule
\end{tabular}
\caption{Main-adapter training summaries read from the saved training runs.
Rows per step equal devices times rows per device.}
\label{tab:training}
\end{table}

\subsection{Representation and feature selection}
\label{app:r62}

At token position $t$ and audited layer $\ell$, let
$h_t^{\rm ada},h_t^{\rm base}\in\R^d$ be the adapted and base model's hidden
states on the same input. Their difference and standardized form are
\begin{equation}
 \delta_t=h_t^{\rm ada}-h_t^{\rm base},\qquad
 q_t=(\delta_t-\mu)\oslash\sigma,
 \label{eq:delta}
\end{equation}
where $\mu,\sigma\in\R^d$ are fitted coordinate means and positive scales,
and $\oslash$ denotes coordinate-wise division. This measures an
adaptation-induced state difference; it is not an additive decomposition of
the model's behavior into independent base and adapter mechanisms.

A sparse autoencoder (SAE) approximates $q_t$ by a few active dictionary
coordinates \citep{cunningham2023sparse,gao2024scaling}:
\begin{equation}
 z_t={\rm TopK}_k\!\left({\rm ReLU}(Eq_t+b_e)\right),\qquad
 \widehat q_t=Dz_t+b_d.
 \label{eq:sae}
\end{equation}
Here $E\in\R^{m\times d}$ is the encoder, $D\in\R^{d\times m}$ the decoder,
$b_e\in\R^m,b_d\in\R^d$ biases, and $m$ the number of features. ReLU replaces
negative values with zero; ${\rm TopK}_k$ retains at most the $k$ largest
positive entries. The encoder and decoder are fitted to minimize
reconstruction error. The principal r4.2 dictionary uses the
layer-26 state, $d=4096$, $m=8192$ and $k=4$. Dictionaries are fixed during
each intervention experiment.

The main dictionary was fitted on evaluation-pool representations. Its
discovery/confirmation split excludes confirmation users from feature
\emph{ranking}, not from dictionary fitting or earlier configuration choices.
We therefore use ``confirmation'' as the name of the fixed split, not as a
claim of untouched end-to-end validation. Independently fitted dictionaries
can use different feature coordinates.

The original ranking compares feature activation on incident tokens with an
insider-free reference. For feature $f\in\{1,\ldots,m\}$, its score is
\begin{equation}
 g_f=\frac{1}{|\mathcal A|}\sum_{(x,t)\in\mathcal A}z_{t,f}(x)
       -\frac{1}{|\mathcal R|}\sum_{(x,t)\in\mathcal R}z_{t,f}(x),
\end{equation}
where $\mathcal A$ is the retained discovery incident-token set and
$\mathcal R$ the retained reference-token set. The five highest-ranked
features form $S$. A natural active control $C$ contains five features with
small ranking gaps and nontrivial activation frequency. This is an activity
screen, not equality of edit norms or edited positions.

Such a contrast can select differences between people rather than differences
between a person's normal and incident days. We tested that possibility in
an exploratory r6.2 branch. In each held-out-user fold, define
$\bar z_f(x)=T^{-1}\sum_{t=1}^{T}z_{t,f}(x)$, the mean over the day's input
token positions. For discovery user $u$, let $\mathcal P_u$ contain its
incident/matched-normal pairs $(i,j)$. Define
\begin{equation}
 d_{u,f}=\frac{1}{|\mathcal P_u|}\sum_{(i,j)\in\mathcal P_u}
           [\bar z_f(i)-\bar z_f(j)],\quad
 D_f=\frac1{|\mathcal U|}\sum_{u\in\mathcal U}d_{u,f},\quad
 W_f=\min_{u\in\mathcal U}d_{u,f},
 \label{eq:ranking}
\end{equation}
where $\mathcal U$ is the set of discovery users. $D_f$ is the equal-user
mean and $W_f$ the weakest-user contrast. Each rule ranks strictly positive
scores; D was the branch's primary rule and W a diagnostic. Comparing a user
with themselves removes an additive, time-invariant identity component if
one is present. It does not cancel arbitrary interactions between profile,
behavior, length and representation.

Changing the reference or balancing users replaces much of the old selection;
token-versus-day weighting changes little. The archived top five fire on
14 of 73 held-out incident days, from two users. D fires on 39 days across all
five users, but a single discovery user supplies 78--94\% of its selected
set's positive contrast. Its held-out incident-minus-normal activation
contrast is positive for three users and negative for two.
W reduces concentration and has positive contrasts for four users, but it
was not causally tested. Activity coverage alone does not establish a shared
incident mechanism.

The same concern applies to r4.2: feature 3843 receives 82.94\% of its
discovery activation mass from RHL0992. The other four selected features
are active for 24--33 of the 35 discovery users. We keep the frozen selection
and disclose the concentration rather than reselecting after outcomes.
These observations do not support an intrinsic profile-versus-behavior
mechanism difference between releases.

\subsection{Interventions, donors and control checks}
\label{app:donors}

\paragraph{Editing a feature set while preserving the reconstruction residual.}
Let $F$ be the selected set $S$ or the active control set $C$, and let $z'_{t,F}$
be the proposed replacement for its coordinates. The requested hidden-state
change is
\begin{equation}
 e_t=\sigma\odot D_{:,F}(z'_{t,F}-z_{t,F}),\qquad
 h'_t=h_t^{\rm ada}+e_t,
\end{equation}
where $D_{:,F}$ contains the decoder columns indexed by $F$ and $\odot$
denotes coordinate-wise multiplication. The original reconstruction residual
is retained. Removing $F$ sets $z'_{t,F}=0$. For a donor day $j$, we first
average its $F$ codes over positions where any member of $F$ is active,
giving a prototype $p_{j,F}$. If none is active, the prototype is zero.
A donor patch at strength $\alpha\in\{1/4,1/2,3/4,1\}$ uses
\begin{equation}
 z'_{t,F}=z_{t,F}+\alpha(p_{j,F}-z_{t,F})
 \quad\text{only where }\sum_{f\in F}z_{t,f}>0.
 \label{eq:patch}
\end{equation}
Other receiver positions are unchanged. The prototype is a within-day
summary, not a semantically aligned token replacement. Each donor is patched
separately; we average its measured loss change, rather than passing one
average-donor state through a nonlinear model.

\paragraph{A signed donor-type contrast.}
Let $L_V(i;F,j,\alpha)$ be the mean prediction loss on target view $V$ of
receiver day $i$ after the patch, and $L_V(i;0)$ its unedited loss. Define
$\Delta_V(i;F,j,\alpha)=L_V(i;F,j,\alpha)-L_V(i;0)$.
For organizational context $c$, let $\mathcal J_b(i,c)$ and
$\mathcal J_a(i,c)$ be the frozen benign-user and incident-day donor sets.
No donor shares the receiver's user. At the primary full dose, define
\begin{align}
 m_V(i;F,k,c)&=\frac1{|\mathcal J_k(i,c)|}
             \sum_{j\in\mathcal J_k(i,c)}\Delta_V(i;F,j,1),\quad k\in\{b,a\},\nonumber\\
 \tau_V(i;c)&=[m_V(i;S,a,c)-m_V(i;S,b,c)]
              -[m_V(i;C,a,c)-m_V(i;C,b,c)].
 \label{eq:tau}
\end{align}
Positive $\tau$ favors benign-donor patches of $S$ relative to incident
donors and the same comparison for $C$. It does not require the benign patch
itself to lower loss; all four components must be inspected. We first average
over available days within each user, then equally over available users.
An absent donor group leaves the paired contrast unavailable.

\begin{table}[t]
\centering
\begin{tabular}{lrr}
\toprule
Behavior-loss contrast & Effect (mn) & 95\% interval \\
\midrule
Removal minus baseline & +4.41 & [+1.81, +7.28] \\
Removal minus matched direction & +5.53 & [+2.89, +8.63] \\
Removal minus active control & +4.28 & [+1.71, +7.18] \\
Incident minus matched-normal contrast & -4.60 & [-9.43, -0.62] \\
\bottomrule
\end{tabular}
\caption{Full-dose r4.2 removal, behavior targets, 35 confirmation users.
Positive removal-minus-baseline means prediction became worse. The negative
last row does not support an incident-specific enhancement under this policy.
Intervals are descriptive 95\% user-resampling intervals.}
\label{tab:ablation}
\end{table}

\paragraph{Donor results.}
The corrected grid contains all 1,362 incident receiver-days, 153,457 donor
pairs and 1,231,742 baseline/edit conditions. Independent aggregation checks
reproduced 920 result cells. The primary behavior-only contrast is negative
in all eight split-by-context cells. Confirmation results appear in
Table~\ref{tab:donors}.

\begin{table}[t]
\centering
\begin{tabular}{lrrrr}
\toprule
Context & Users & $\tau$ & 95\% interval & Users with $\tau_u<0$ \\
\midrule
Team & 32/35 & -1.41 & [-2.33, -0.58] & 23/32 \\
Role & 32/35 & -1.86 & [-2.80, -0.99] & 25/32 \\
Department & 34/35 & -2.05 & [-2.93, -1.21] & 25/34 \\
Dept.\ $\times$ role & 32/35 & -1.75 & [-2.60, -0.97] & 28/32 \\
\bottomrule
\end{tabular}
\caption{Full-dose mean-donor behavior contrast on the r4.2 confirmation split.
Effects are in millinats per target token (mn; $1\,\mathrm{mn}=10^{-3}$ nats).
Intervals resample receiver users with the dictionary, feature selection and
donor bank fixed. Different contexts cover different available users.}
\label{tab:donors}
\end{table}
\begin{figure}[t]
\centering
\includegraphics[width=\linewidth]{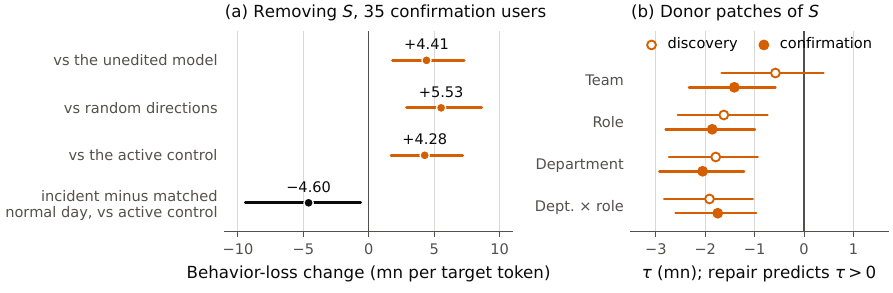}
\caption{The removal and donor results of Tables~\ref{tab:ablation} and~\ref{tab:donors}, with the donor contrast's
discovery split. (a) Full-dose removal of $S$, behavior targets, 35 confirmation users: removal raises behavior loss
against each comparison, but its incident-minus-matched-normal contrast against the active control is negative.
(b) The donor contrast $\tau$ in each matching context: its point estimate is negative in every context and split,
where the repair reading predicts $\tau>0$. Bars are descriptive 95\% user-resampling intervals.}
\label{fig:interventions}
\end{figure}

The interpretation remains conditional. Natural active controls are not
norm matched, and selected benign/incident donor edits differ in size.
Paired contrasts exclude users lacking another insider in their context.
On 123 days the selected set is inactive, so its edit is zero while controls
can still act; these days remain in the primary estimate. The dictionary
and prior configuration choices saw evaluation users. These limitations
permit a descriptive joint-set effect, not a claim that the features encode
an insider concept or that benign patches repair behavior.

The principal r4.2 selected feature IDs are
$S=\{3843,\allowbreak 3230,\allowbreak 3199,\allowbreak 3975,\allowbreak 1394\}$ and active controls
are $C=\{1087,\allowbreak 3632,\allowbreak 1319,\allowbreak 3127,\allowbreak 72\}$. IDs refer to the single fixed layer-26,
8,192-feature dictionary. They have no shared coordinate meaning in an
independently fitted dictionary. All four nonzero strengths and all planned
donor conditions were saved before aggregation.

The donor policy first uses exact organizational context, then its declared
primary-field and global fallbacks, with same-user exclusion after fallback.
An empty post-exclusion pool remains empty. In the accepted grid, every
incident-donor pair is exact-context. A user lacks a paired contrast precisely
when no other insider shares the relevant context. Benign donors come from
official non-insider identities; incident donors are corrected incident days
of other users. The frozen equal-budget subset is used only for the separate
searched-best comparison, not substituted for the primary full-pool mean.

Let $\mathcal R_u(c)$ be the days of user $u$ for which both donor groups
and target view are available, and $\mathcal U_g(c)$ the users in split $g$
with at least one such day. The reported mean is
\begin{equation}
 \tau_{V,g,c}=\frac{1}{|\mathcal U_g(c)|}\sum_{u\in\mathcal U_g(c)}
      \frac{1}{|\mathcal R_u(c)|}\sum_{i\in\mathcal R_u(c)}\tau_V(i;c).
\end{equation}
The all-planned-user endpoint remains unavailable if any required day is
missing. No missing donor group is filled with zero. All four component
means below use exactly the same receivers as their paired contrast; averages
on each arm's own available population need not reconstruct $\tau$.

For available-case endpoints, the 95\% intervals use 10,000 receiver-user
bootstrap draws. The donor bank, dictionary, selection and scored
interventions are fixed. These are not claims of frequentist coverage for
new users or new donors. A leave-one-user-out sensitivity reports changes
in point estimates, not independent replications. Shared donor users can
induce dependence when the target instead averages over new donor banks;
alternative crossed/dyadic calculations examined during development do not
yet certify that broader inference.

An offline fidelity audit covers 1,742 r4.2 receiver-days: the 1,362 incident
days and 380 matched normal days. It reconstructs native edited states and
re-encodes them with the same SAE on CPU. Unedited active-feature identities
agree with the archived GPU codes on all 476,429 tokens; values agree within
tolerance, not bitwise.

Removal eliminates all 15,833 originally active entries of $S$. Previously
inactive members of $S$ nevertheless activate on 127 tokens, carrying 0.43\%
of the original mass. A norm-matched random dictionary-direction control
retains 97.30\% of original $S$ mass pooled, but 570 tokens have retention
below $10^{-6}$ and 7,499 show amplification. The separate active-feature
control mostly preserves $S$ while editing different positions: only
68 of its 7,394 edited tokens overlap the 15,775 S-active tokens.

These are different control policies. One matches size at S-active positions;
the other uses naturally active alternative features. Neither check
establishes that all nuisance effects are matched. The offline audit does
not measure downstream loss anew or certify post-edit CPU/GPU code agreement.

\subsection{Evidence provenance}
\label{app:scope}

Generative AI assisted the analysis code, the experiment runs and the drafting of this paper; the Generative AI
Usage Statement describes how.
The companion evidence snapshot binds machine-readable reviewed summaries
by SHA-256. The exhibit generator verifies those hashes and the donor-result
receipt before formatting tables. Source identifiers, the retained-claim map
and reproduction instructions are kept with the LaTeX source in the paper's repository. This packages the accepted
summary evidence; it does not imply that every raw GPU array is distributed
with the manuscript.

The label repair corrects calendar membership, receiver coverage and donor
eligibility. It does not require retraining the main benign-user adapters;
their training membership was checked separately. Where a later serializer
or adapter differs, it is treated as a distinct experiment. Numerical replay
supports implementation consistency on checked configurations, not hardware
independence for all future interventions.

\FloatBarrier\section{The control's metric: definitions and tests}
\label{app:geometry}

\paragraph{The preserving control.}
At a targeted token $t$ (a position where the removal request is nonzero), let $e_t$ be the realized removal edit and
$\rho_t=\|e_t\|_2$ its size. Let $C_t$ stack the encoder rows, divided by $\sigma$, of the selected features and of the
features active in the unedited state that are not selected. Let $c_{0,t}$ satisfy the reader equations: the selected
readers keep their unedited values, and the other active readers take their values under the removal. For a positive
diagonal metric $M$, the control solves
\begin{equation}
 v_t=\arg\min_v\,(v-e_t)^\top M(v-e_t)\quad\text{subject to}\quad C_tv=C_tc_{0,t},\ \ \|v\|_2\le\rho_t.
\end{equation}
The raw geometry uses $M=I$; the SAE geometry uses $M={\rm diag}(1/w)$ with $w\propto\sigma^2$. Tokens at which a
selected feature's top-$k$ membership could change are either solved with certified support constraints or reported
without a control. Every control is refined until its BF16 realization passes the reader tolerances, and is no larger
than $1.005\rho_t$. Each receiver also receives 24 perturbed controls per geometry, each a small random displacement
within the reader null space that must pass the same gates. A receiver is admitted only if both geometries pass for
the anchor and all 48 perturbations. Its contrast $\theta$ is the removal's behavior loss minus the mean over the 24
perturbed controls, with a standard error from their spread.

\paragraph{Distance between metrics, and the radii.}
For $M={\rm diag}(m)$ and $M'={\rm diag}(m')$ with $m'_j=m_je^{\eta_j}$, the Hilbert projective distance is
$d_H=\max_j\eta_j-\min_j\eta_j$. Rescaling $M$ leaves the control unchanged, so the rate vector
$c=\partial\theta/\partial\eta$ at $\eta=0$ sums to zero. The largest first-order change within $d_H\le r$ is then
$(r/2)\|c\|_1$, attained at $\eta_j=(r/2)\,{\rm sign}(c_j)$, which gives Equation~\ref{eq:rstar}. For random signs,
$\eta_j=(r/2)\varepsilon_j$, Hoeffding's inequality (two-sided, and so conservative for erasure, which is one-sided) bounds the probability that
the first-order change reaches $|\theta|$ by 0.05 for all $r\le r_{0.05}=2|\theta|/(z\|c\|_2)$ with $z=\sqrt{2\ln 40}$. Hence
$r_{0.05}/r^\ast=(\|c\|_1/\|c\|_2)/z$, a property of the rate vector's spread. The two declared geometries are
$d_H=\CtlEightCap$ apart on the 8B and \CtlThreeCap{} apart on the 3B. Test radii are capped at that distance.

\begin{proposition}\label{prop:gap}
Let $\theta(\eta)$ be differentiable at $\eta=0$ with gradient $c\in\R^d$, and invariant to $\eta\mapsto\eta+a\mathbf 1$,
so that $\mathbf 1^\top c=0$. For $r>0$:
\begin{enumerate}
 \item $\sup\{c^\top\eta:\ \max_j\eta_j-\min_j\eta_j\le r\}=(r/2)\|c\|_1$;
 \item if $\eta=(r/2)\varepsilon$ with independent random signs $\varepsilon_j=\pm1$, then $c^\top\eta$ has mean $0$ and
 standard deviation $(r/2)\|c\|_2$, and $\Pr(|c^\top\eta|\ge t)\le2\exp\{-2t^2/(r^2\|c\|_2^2)\}$;
 \item to first order, the outcome-aware change erases $\theta$ at $r^\ast=2|\theta|/\|c\|_1$, while random changes
 erase it with probability at most $0.05$ for every $r\le r_{0.05}=r^\ast(\|c\|_1/\|c\|_2)/z$, $z=\sqrt{2\ln40}$.
\end{enumerate}
For $c\neq0$ the ratio $\|c\|_1/\|c\|_2=\sqrt{d_{\rm eff}}$ lies in $[\sqrt2,\sqrt d]$; the upper end requires all
$|c_j|$ equal, which $\mathbf 1^\top c=0$ allows only for even $d$.
\end{proposition}
\begin{proof}
(1) Write $\eta=a\mathbf 1+\delta$ with $a=(\max_j\eta_j+\min_j\eta_j)/2$, so that $|\delta_j|\le r/2$. Then
$c^\top\eta=c^\top\delta\le(r/2)\|c\|_1$, with equality at $\delta=(r/2)\,{\rm sign}(c)$. (2) is Hoeffding's
inequality for $\sum_j\varepsilon_j(r/2)c_j$. (3) Solve $(r/2)\|c\|_1=|\theta|$, and set the bound in (2) to $0.05$ at
$t=|\theta|$. The upper end of the ratio's range is the Cauchy--Schwarz inequality. For the lower end, the positive
and negative parts of $c$ each sum to $\|c\|_1/2$ because $\mathbf 1^\top c=0$, so
$\|c\|_2^2\le(\|c\|_1/2)^2+(\|c\|_1/2)^2$.
\end{proof}
The invariance premise holds in every run. All \CtlPremiseCount{} receiver and population rate vectors of the audit
runs sum to zero to within \CtlPremiseBound{} of their $\ell_1$ norms, and so do all \CtlPremisePublicCount{} nonzero
document and population rate vectors of the public models (within \CtlPremisePublicBound).
Popoviciu's inequality bounds the size of any direction at distance $r$, not its alignment with $c$: a direction whose
coordinates span $r$ has root-mean-square at most $r/2$ after centering, the value random signs attain. A structured
outcome-blind direction can therefore still move $\theta$ more than random signs do if it happens to be correlated
with $c$. The obvious structured alternative, moving toward the other declared metric, moves it by only
\CtlAblLmPathMin\% to \CtlAblLmPathMax\% at the radius (Section~\ref{sec:sensitivity}).
Table~\ref{tab:sensitivity} gives these quantities by setting. First order also describes the measured random changes:
of the \CtlAblNatN{} in the public-model and autoencoder runs of Table~\ref{tab:sensitivity}, \CtlAblNatOut{} fall outside
its 95\% range (about \CtlAblNatExp{} expected), and their median is \CtlAblNatMed{} standard deviations
(\CtlAblNatMedExp{} expected).

\begin{table}[t]
\centering
{\footnotesize\setlength{\tabcolsep}{4pt}\begin{tabular}{@{}lrrrrrrr@{}}
\toprule
 & & & \multicolumn{2}{c}{Random change (\% of $\theta$)} & \multicolumn{2}{c}{$r^\ast$/gap (\%)} & \\
\cmidrule(lr){4-5}\cmidrule(lr){6-7}
Setting & Pairs & $d_{\mathrm{eff}}$ & Random SD & Largest native & Hilbert & $\ell_2$ & Path \\
\midrule
Web text & 22 & 199--1,118 & 3.0--7.1 & 1.4--16.7 & 2.0--17.6 & 12.1--141.4 & $-$5.0 to $+$7.8 \\
Security logs & 36 & 254--970 & 3.2--6.3 & 1.2--11.0 & 1.6--6.7 & 6.5--31.7 & $-$1.7 to $+$2.4 \\
CERT detectors (Qwen)$^{\ast}$ & 8 & 753--1,804 & 2.4--3.6 & 1.5--17.4 & 4.8--15.0 & 15.8--54.0 & $-$0.3 to $+$4.5 \\
Autoencoder (width 64) & 4 & 17--27 & 19.1--23.9 & 9.3--34.7 & 16.6--73.0 & 21.2--89.3 & $-$16.8 to $+$19.6 \\
\bottomrule
\end{tabular}}
\caption{The test's own choices, per setting, at the population radius (post hoc, first order unless stated).
$d_{\mathrm{eff}}$: effective number of coordinates of the rate vector. Random SD: the first-order standard deviation
of the change from a random-sign metric at the radius, $1/\sqrt{d_{\mathrm{eff}}}$ of $\theta$ (over
\CtlAblDraws{} draws per pair, none of which erases $\theta$). Largest native: the larger of the run's two measured
random changes. $r^\ast$/gap: the erasing metric's distance from the declared one as a percentage of the distance
between the two standard metrics, on the Hilbert scale and in $\ell_2$ on the log-weights, both modulo a common scale.
Path: the change in $\theta$, to first order, from moving the same distance toward the other standard metric. The rows
include the web-text and security-log replications; the public models reading CERT days are not covered.
$^{\ast}$Hilbert, $\ell_2$ and path: six of eight pairs, since the 8B's SAE scale vector was not at hand.}
\label{tab:sensitivity}
\end{table}

\paragraph{Per-receiver tests.}
Each test receiver gets five labels, with $r_{\rm test}=\min(r^\ast,d_H({\rm raw},{\rm SAE}))$. M0 rebuilds the declared
control. $A_{1/2}$ and $A_1$ move the metric to distance $r_{\rm test}/2$ and $r_{\rm test}$ along
$-{\rm sign}(\theta)\,{\rm sign}(c)$. $U_1$ and $U_2$ use random signs at distance $r_{\rm test}$. For each label,
$D=\Delta(\text{label})-\Delta(M0)$ is compared with its first-order prediction $\pi$; $\sigma_U$ is the first-order
scale of a random-sign change. The rules are:
\begin{itemize}
 \item P1: $|D(A_{1/2})-\pi_{1/2}|\le0.25|\pi_{1/2}|+2{\rm SE}$;
 \item P2: $|D(A_1)-\pi_1|\le0.5|\pi_1|+2{\rm SE}$;
 \item P3: $|D(U_k)|\le3\sigma_U+2{\rm SE}$;
 \item P4: $|D(A_1)|>\max_k|D(U_k)|$.
\end{itemize}
A receiver is informative if all five labels are admitted and $|\pi_{1/2}|\ge2\,{\rm SE}_D$. A decision needs at least
six informative receivers, and a property holds if it holds on at least 75\% of them. The decision is
\emph{validated} if P1 and P2 hold, and \emph{curvature-limited} if only P1 holds. Among the 8B's fresh receivers
(exploratory), \CtlEightCrossedRaw{} and \CtlEightCrossedSae{} reversed by $r^\ast$, at a median
\CtlEightNativeRaw--\CtlEightNativeSae{} $r^\ast$ among those that did.

\paragraph{Population tests.}
The equal-user mean contrast $\theta_{\rm pop}$ and rate vector $c_{\rm pop}$ are taken over every admitted receiver,
and $r_{\rm pop}=\min(2|\theta_{\rm pop}|/\|c_{\rm pop}\|_1,\ d_H({\rm raw},{\rm SAE}))$. One receiver per user is then
scored under three shared metrics:
\begin{itemize}
 \item $A_{\rm pop}$, along $-{\rm sign}(\theta_{\rm pop})\,{\rm sign}(c_{\rm pop})$ at distance $r_{\rm pop}$;
 \item $U_{{\rm pop}1}$ and $U_{{\rm pop}2}$, with random signs at the same distance.
\end{itemize}
P$_{\rm pop}$1 requires $|\bar D(A_{\rm pop})-\pi_P|\le0.5|\pi_P|+2{\rm SE}$, and P$_{\rm pop}$2 requires
$|\bar D(U)|\le3\sigma_P+2{\rm SE}$. Both are computed over P$^\ast$, the users all of whose labels were constructed,
and need $|{\rm P}^\ast|\ge25$. SE is the measurement error of the mean native change, with the users fixed. The 3B
runs add the rule $|\pi_P|\ge4\,{\rm SE}$, which asks whether the predicted change is visible but not whether $\theta_P$
itself differs from zero. The native contrast under $A_{\rm pop}$ is always reported.

\paragraph{Details behind Section~\ref{sec:results}.}
The contrasts were distinguishable from zero (post hoc): \CtlStarZMin--\CtlStarZMax{} standard errors across the users
of the audit detectors' population samples, \CtlUnitZWebMin--\CtlUnitZWebMax{} across documents on web text, and
\CtlUnitZLogsMin--\CtlUnitZLogsMax{} and \CtlUnitZCertMin--\CtlUnitZCertMax{} with documents clustered by user on the
security logs and the CERT days; the 3B's seeds and the Gemma-3-1B detector are the exceptions
Section~\ref{sec:res-blind} names. The Gemma-3-1B detector's contrast lies \CtlResXSixGemmaThetaTRaw{} and
\CtlResXSixGemmaThetaTSae{} standard errors from zero across its users, and its random bounds, like the seeds', are weak
or undefined (the Gemma detector's reach \CtlResXSixGemmaRandLow\%), though their largest random moves stayed within the
declared bounds. Per receiver, the declared first-order decision was validated in \CtlValidatedTests{} of the six
model--geometry tests; the sixth, the 8B's raw test, was \CtlEightDecisionRaw. Native changes ran larger than first order
in all detector pairs but four: the SAE-scaled metric of seeds 43 and 45 and of the LANL detector, and the
Gemma-3-1B detector's raw metric,
where the native change lay within a standard error of zero. At least 0.5\% of the contrast remained at $r^\ast$
(Table~\ref{tab:runs}) in
DeepSeek, whose native changes fell slightly short of first order, in Gemma's LANL run and its two replications at the
raw metric, in both replications of Llama's TWOS run, in the 3B's cohort 2, seeds 44 and 45 and seed~43's raw metric,
and in the Gemma-3-1B detector, where the radius is set on every admitted receiver while the contrast is measured on the
users all of whose labels were constructed, and on those users first order at $r^\ast$ predicts less than the whole
contrast. The public-model rules' requirement of erasure within the gap is met trivially ($r^\ast$ is at most
\CtlSetGapMax\% of the gap); the detectors' rules have none.

\FloatBarrier\section{The settings in brief}

\subsection{Public models on web text}
\label{app:public}

Public settings test whether the pattern depends on the audited detectors. All use OpenWebText documents
\citep{gokaslan2019openwebtext}, split by a seeded permutation into 200 selection, 60 pilot and 1,000 confirmation
documents (the protocol for Pythia, Gemma and Llama uses no pilot documents), and each declares its predictions before its confirmation
documents are run. No CERT data is involved.

\paragraph{GPT-2 small.}
\begin{itemize}
 \item \textbf{Model and SAE:} GPT-2 small \citep{radford2019language}, with a public residual-stream SAE at the input
 to block 8 \citep{bloom2024gpt2saes} (24,576 ReLU features, MIT licence).
 \item \textbf{Input convention:} the SAE reads the residual stream centered across coordinates. A positive control
 confirmed it: the fraction of variance left unexplained is \CtlPMFVUCentered{} with centering and \CtlPMFVURaw{}
 without.
 \item \textbf{Documents:} an end-of-text token plus each document's first 127 tokens.
 \item \textbf{Selected set:} the five features most active on digit tokens relative to other tokens, chosen on the
 selection documents.
 \item \textbf{Declared metrics:} raw, and inverse squared per-coordinate scale, \CtlPMCap{} Hilbert units apart.
 \item \textbf{Construction:} removal and control as in Appendix~\ref{app:geometry}, within the centered subspace.
 ReLU codes need no top-$k$ support constraint. The model runs in float64.
\end{itemize}

\paragraph{DeepSeek-R1-Distill-Qwen-1.5B.}
\begin{itemize}
 \item \textbf{Model and SAE:} DeepSeek-R1-Distill-Qwen-1.5B \citep{deepseek2025r1}, with EleutherAI's TopK SAE on the
 layer-14 MLP output (65,536 latents, $k=32$, trained on FineWeb-Edu text) \citep{eleuther2025deepseeksae}. On the
 selection documents it leaves \CtlPMTwoFVU{} of the MLP output's variance unexplained, against \CtlPMTwoFVUResid{}
 when applied to the residual stream.
 \item \textbf{Documents:} each document's first 128 tokens (this tokenizer adds no beginning-of-sequence token).
 \item \textbf{Construction:} edits are added to the MLP output. At the control state every selected feature must keep
 its top-$k$ membership with a margin of $10^{-3}$; otherwise the document is not constructed. The model runs in
 float32 with float64 loss accumulation and deterministic kernels, and the declared control is rebuilt exactly.
 \item \textbf{Declared metrics:} raw, and inverse squared per-coordinate scale, \CtlPMTwoCap{} Hilbert units apart.
 \item \textbf{Selection, and the order of events:} the rule changed twice on the pilot documents. GPT-2's digit rule
 picked a feature active on most tokens. Restricted to selective features, it picked features with no population
 effect: $\theta_P$ lay within \CtlPMTwoVtwoZ{} standard errors of zero, yet its predicted change, at
 \CtlPMTwoVtwoInfMin{} and \CtlPMTwoVtwoInfMax{} standard errors, met the size part of GPT-2's informativeness rule. We then selected the five features whose removal most increases the loss to
 first order, among features active on 0.1--1\% of tokens, and added the requirement $|\theta_P|\ge4\,{\rm SE}(\theta_P)$.
 Both changes and the predictions were committed before the confirmation documents were run.
\end{itemize}

\paragraph{Protocol and verdicts.}
\begin{itemize}
 \item Three predictions per metric were declared before the confirmation documents:
 \begin{itemize}
  \item the population test P$_{\rm pop}$1 holds for the outcome-aware shared metric;
  \item P$_{\rm pop}$2 holds for two random-sign shared metrics;
  \item $G\ge5$.
 \end{itemize}
 \item They count as informative if $|{\rm P}^\ast|\ge100$ and $|\pi_P|\ge4\,{\rm SE}$, with SE the between-document
 standard error. That rule does not require the population contrast itself to differ from zero, which DeepSeek's
 declaration adds. Checked afterwards, GPT-2's contrast lies \CtlPMThetaZRaw{} and \CtlPMThetaZSae{} standard errors
 from zero (bootstrap 95\% intervals \CtlPMBootLoRaw--\CtlPMBootHiRaw{} and \CtlPMBootLoSae--\CtlPMBootHiSae\,mn).
 \item Of the 1,000 confirmation documents, \CtlPMAdmitted{} (GPT-2) and \CtlPMTwoAdmitted{} (DeepSeek) had targeted
 positions and were admitted.
 \item Both declared verdicts are that the pattern replicates: all three predictions are informative and hold at both
 metrics. Second implementations agreed on all \CtlPMChecks{} and all \CtlPMTwoChecks{} comparisons.
\end{itemize}

\paragraph{Pythia-410m: a declared run stopped by a numerical gate, and a post hoc rerun.}
\begin{itemize}
 \item \textbf{Model and SAE:} Pythia-410m \citep{biderman2023pythia}, with EleutherAI's TopK SAE on the layer-12 MLP
 output (65,536 latents, $k=32$) \citep{eleuther2024pythiasae}. It leaves \CtlPyFVU{} of the MLP output's variance
 unexplained, against \CtlPyFVUWrong{} on the residual stream.
 \item \textbf{Construction:} as for DeepSeek, except that every reader of Pythia's residual stream sits behind a
 LayerNorm, so edits live in the centered subspace, as for GPT-2. Documents are each document's first 128 tokens (no
 beginning-of-sequence token). The declared metrics are \CtlPyCap{} Hilbert units apart, a cost ratio of
 \CtlPyCapCost$\times$.
 \item \textbf{Protocol:} declared before the model was downloaded, together with Gemma-3-1B and Llama-3.2-1B. The
 selection rule, metrics and predictions are DeepSeek's, without pilots, plus two
 riskier predictions: the outcome-aware metric erases or reverses the contrast
 (${\rm sign}(\theta_P)\,\theta_P^{A}\le2\,{\rm SE}$), and it lies closer to the declared metric than the two declared
 metrics lie to each other. A regression rebuilt DeepSeek's committed records exactly with the generalized code, and two
 deliberately wrong versions (an off-by-one layer and a missing decoder-bias shift) each produced at least
 \CtlPMThreeRegWrong{} mismatches. Both passed the first-order gate, which therefore does not check the hookpoint.
 \item \textbf{Declared run:} float32, as for DeepSeek. Of 1,000 documents \CtlPyAdmittedFp{} were admitted, then the
 first-order gate stopped the run: \CtlPyGatePass{} of 6 cases passed and one missed at \CtlPyGateMiss{} times the
 prediction. The declared verdict is \emph{does not replicate}.
 \item \textbf{Diagnosis (post hoc):} recomputed in float64, the six ratios are \CtlPyDiagLo--\CtlPyDiagHi{} at the
 gate's radius and within \CtlPyDiagLast{} of 1 at an eighth of it, while in float32 they scatter from
 \CtlPyNoiseLo{} to \CtlPyNoiseHi{} as the radius shrinks. That is rounding noise in the forward pass on predicted
 changes of \CtlPyPredLo--\CtlPyPredHi\,mn, not a failure of first order. DeepSeek needed float32 because float64 did
 not fit; Pythia does not, and our protocol should not have copied the choice.
 \item \textbf{Rerun (post hoc):} float64, with the same selected features, documents, code path and predictions,
 declared as an amendment before it ran. It replicates (Table~\ref{tab:runs}): a reweighting of
 \CtlPyDistMin--\CtlPyDistMax$\times$ erases or reverses the contrast, and random ones move it by at most
 \CtlPyBlindMax\%. Its declared metrics disagree about the contrast's size by a factor of \CtlPyOtherFactor{}, though
 not about its sign. We report it beside the declared verdict, not in place of it.
\end{itemize}

\paragraph{Gemma-3-1B and Llama-3.2-1B: declared runs in float64.}
\begin{itemize}
 \item \textbf{Models and SAEs:} Gemma-3-1B, pretrained \citep{gemmateam2025gemma3}, with a Gemma Scope 2 JumpReLU SAE
 on the residual stream after layer 13 of 26 (16,384 latents; the release's 50\%-depth layer)
 \citep{google2025gemmascope2}; and Llama-3.2-1B, from the Llama 3 family \citep{grattafiori2024llama3}, with
 EleutherAI's TopK SAE on the layer-8 MLP output (131,072 latents, $k=32$) \citep{eleuther2025llamasae}. The Gemma SAE
 leaves \CtlGmFVU{} of its site's variance unexplained, against \CtlGmFVUWrong{} on the residual stream entering layer
 13; the Llama SAE leaves \CtlLlFVU{}, against \CtlLlFVUWrong{} on the residual stream.
 \item \textbf{Construction:} as for DeepSeek. Both models use RMSNorm, so edits are not centered. Gemma's JumpReLU codes
 are elementwise, so instead of a TopK membership check its selected codes are recomputed at each control and must
 equal their base values. Both tokenizers add a beginning-of-sequence token, which edits never touch. The declared
 metrics are \CtlGmCap{} (Gemma) and \CtlLlCap{} (Llama) Hilbert units apart, cost ratios of \CtlGmCapCost$\times$
 and \CtlLlCapCost$\times$.
 \item \textbf{Precision:} after Pythia, both were to run in float64. Before either ran, we found that the installed
 library computes these models' RMSNorm layers in float32 even in a float64 model, rounding every norm's input. An
 amendment written before any stage on either model computes them in float64 (\CtlGmPromoted{} and
 \CtlLlPromoted{} norm modules), checks each against a float64 reference at every load, and adds a precision probe on
 selection documents: along the loss gradient, the native change matches first order within $\CtlDecProbeMax$ at
 $10^{-5}$\,nats, against a declared bound of 1\%. The amended code still rebuilt DeepSeek's and Pythia's float64
 records exactly; the same Pythia entry in float32 differed in \CtlAtwoRegFpWrong{} quantities on
 \CtlAtwoRegDocs{} documents.
 \item \textbf{Declared runs:} Gemma admitted \CtlGmAdmitted{} of 1,000 documents (\CtlGmNoop{} had no targeted
 position) and Llama \CtlLlAdmitted{} (\CtlLlNoop{} with no targeted position, \CtlLlNotCon{} without a feasible
 control). Every first-order gate case passed (native over predicted \CtlGmGateLo--\CtlGmGateHi{} and
 \CtlLlGateLo--\CtlLlGateHi). Both replicate at both metrics (Table~\ref{tab:runs}): Gemma's contrast reverses,
 and Llama's is erased without reversing.
\end{itemize}

\paragraph{Replications on fresh web text.}
Gemma's, Llama's and Pythia's (float64) runs were rerun on two further disjoint blocks of web text. All six blocks
replicated under the web-text rule, and at least 80\% of the contrast was removed (one-sided 95\%) in
\CtlWebRepEighty{} of \CtlWebRepEightyN{} block--metric pairs. The declared reading requires that at both metrics in
both blocks, so it held for Llama and Pythia but not for Gemma, whose second block at the raw metric reversed the
contrast by its point estimate (\CtlWebRepGemmaLeft\% left) but missed the bound by \CtlWebRepGemmaMiss\,mn. In one
block a random metric moved the contrast by \CtlWebRepRandMax\%, beyond first order's 95\% range of
\CtlWebRepRandQ\%.

\subsection{Public models reading security logs}
\label{app:logs}

Does the pattern of Section~\ref{sec:geometry} hold when the documents are real security logs rather than web text? A
fourth protocol, fixed before any document was built, answers this for the three models declared last
(Appendix~\ref{app:public}), with the same SAEs, sites and float64 precision, and the same code extended to read the
log corpora; Pythia's float64 run is declared here. The models are general-purpose language models reading serialized
logs, not the detectors of the main text, and their SAEs are the public releases used on web text, not refit to logs.
\begin{itemize}
 \item \textbf{TWOS} \citep{harilal2017twos} records 24 participants, in six teams of four, over a five-day competition
 that incentivized both normal and malicious behavior, including masquerade and traitor attacks. We use this project's
 earlier serialization of their activity into 10-minute windows. A document is four consecutive windows of one user's
 day: the first in full, then the session lines of the next three. \CtlSecDocsTwosMin{} documents reach 128 tokens
 in every tokenizer.
 \item \textbf{LANL} \citep{kent2016cybersecurity,kent2015comprehensive} holds 58 days of de-identified events from the
 laboratory's network. We take the first 20 million authentication events, keep human users, cut each user's day into
 windows of 32 consecutive events serialized in order, and draw 10,000 windows with a fixed seed;
 \CtlSecDocsLanlMin--\CtlSecDocsLanlMax{} reach 128 tokens, depending on the tokenizer.
 \item \textbf{Protocol.} As before, a document is its first 128 tokens and a fixed permutation assigns 200 selection
 and 1,000 confirmation documents. For each run, the selection stage (no outcome) ran first, and a declaration rendered
 from its records was committed before the confirmation ran. Every hookpoint control passed: the SAEs leave
 \CtlSecFVUMin--\CtlSecFVUMax{} of their site's variance unexplained on the logs, against
 \CtlSecFVUWrongMin--\CtlSecFVUWrongMax{} at the wrong site. Predictions, verdict rules and the split-sample test are
 those of the three models declared together.
 \item \textbf{Replications.} Each pair ran twice more, on the next two blocks of the same permutation, each again
 200 selection, 60 unused and 1,000 confirmation documents; a pair's three blocks share no document. Each replication
 repeated the selection stage on its own block, and a declaration rendered from those records was committed before any
 replication's confirmation ran. The split-sample test was not run for the replications.
 \item \textbf{Results.} Each of the eighteen runs admitted \CtlSecAdmittedMin--\CtlSecAdmittedMax{} of its 1,000
 documents, and every first-order gate case passed (native over predicted \CtlSecGateLo--\CtlSecGateHi). All six
 declared runs replicate at both metrics, as do \CtlSecRepYes{} of the \CtlSecRepN{} replications
 (Table~\ref{tab:runs}); in each run that replicates, the outcome-aware metric erases the contrast without reversing
 it. The other two replications (\CtlSecRepNoList) each fail only the erasure prediction: there the native change
 is \CtlSecRepLeftRatioMin--\CtlSecRepLeftRatioMax{} times first order, which leaves
 \CtlSecRepLeftMin--\CtlSecRepLeftMax\% of the contrast, \CtlSecRepLeftZMin--\CtlSecRepLeftZMax{} standard errors
 from zero (\CtlSecRepLeftClusterTMin--\CtlSecRepLeftClusterTMax{} with documents clustered by user). In the split-sample
 test the half-built metric removed \CtlSecXsRemovedMin--\CtlSecXsRemovedMax\% of the other half's contrast in all
 \CtlSecXsDirections{} directions; \CtlSecXsHold{} runs held and \CtlSecXsFail{} failed the declared rule, each on one
 direction whose remainder (\CtlSecXsLeftMin\% or \CtlSecXsLeftMax\%) was detectable only because its contrast lay
 \CtlSecXsLeftZMin--\CtlSecXsLeftZMax{} standard errors from zero between documents. Across the eighteen runs native
 changes are
 \CtlSecRatioMin--\CtlSecRatioMax{} times first order, and $G$ is \CtlSecGMin--\CtlSecGMax. The declared metrics
 disagree about the contrast's size by up to \CtlSecOtherMax\%, never about its sign.
\end{itemize}

\subsection{Training seeds of the 3B detector}
\label{app:xfive}

Each detector in Table~\ref{tab:runs} is one training run. A declared replication retrains the Qwen2.5-3B r4.2
adapter with seeds 43, 44 and 45, by seed 42's command with only the seed changed, and makes each seed's own
layer-24 adaptation-delta SAE and top-five feature set by seed 42's commands. Each seed then runs cohort 2's two
declared tests on the same 600 receiver-days (users 241--440, three training-split days each): the radius test,
and the population predictions under the first cohort's rules. A seed replicates if the radius test is supported and
both population predictions hold at both metrics; anything else is reported as found. Seed~43 was trained in August
2026, before any of this work; seeds 44 and 45 were trained for this replication on a different GPU type with the same script,
configuration, data and base model, checked by hash.
Seed~43 admitted \CtlXFiveAdmitted{} of the \CtlXFiveOf{} receivers (\CtlXFiveSeedFortyTwo{} at seed~42), leaving
\CtlXFiveHOneRaw{} and \CtlXFiveHOneSae{} receivers with contrasts of at least 1\,mn, against the radius test's
declared minimum of 10, so that test was \CtlXFiveXFourVerdict{} and seed~43 \CtlXFiveVerdict. Its population sample
had \CtlXFivePstarRaw{} users at the raw metric and \CtlXFivePstarSae{} at the SAE-scaled one; its population predictions held
at both metrics with its users fixed, but at the raw metric the predicted change was \CtlXFiveFalsifRaw{} of the declared
four standard errors, so that prediction was not informative. Its native population
change was \CtlXFiveRatioRaw{} and \CtlXFiveRatioSae{} times first order, and its largest random change
\CtlXFiveRandomRaw\% and \CtlXFiveRandomSae\%.
Seed~44 admitted \CtlXFiveFortyFourAdmitted{} receivers, of which \CtlXFiveFortyFourHOneRaw{} and \CtlXFiveFortyFourHOneSae{}
had contrasts of at least 1\,mn, with a median per-receiver $r^\ast$ of \CtlXFiveFortyFourHOneMedRaw{} and
\CtlXFiveFortyFourHOneMedSae{}; its radius test was \CtlXFiveFortyFourXFourVerdict, and with both population predictions
holding at both metrics seed~44 \CtlXFiveFortyFourVerdict. Its population sample had \CtlXFiveFortyFourPstarRaw{} and
\CtlXFiveFortyFourPstarSae{} users, its native population change was \CtlXFiveFortyFourRatioRaw{} and
\CtlXFiveFortyFourRatioSae{} times first order, and its largest random change \CtlXFiveFortyFourRandomRaw\% and
\CtlXFiveFortyFourRandomSae\%.
Seed~45 admitted \CtlXFiveFortyFiveAdmitted{} receivers, of which only \CtlXFiveFortyFiveHOneRaw{} and
\CtlXFiveFortyFiveHOneSae{} reached 1\,mn, so its radius test was \CtlXFiveFortyFiveXFourVerdict{} (median per-receiver
$r^\ast$ \CtlXFiveFortyFiveHOneMedRaw{} and \CtlXFiveFortyFiveHOneMedSae, descriptive only) and seed~45
\CtlXFiveFortyFiveVerdict, although both population predictions held at both metrics. Its population sample had
\CtlXFiveFortyFivePstarRaw{} and \CtlXFiveFortyFivePstarSae{} users, its native population change was
\CtlXFiveFortyFiveRatioRaw{} and \CtlXFiveFortyFiveRatioSae{} times first order, and its largest random change
\CtlXFiveFortyFiveRandomRaw\% and \CtlXFiveFortyFiveRandomSae\%.
In all three seeds the radius test's other three predictions held at both metrics, and the first cohort's combined rule
read \emph{inconsistent} at both metrics because its per-receiver prediction failed, as at seed~42. So
\CtlXFiveSeedsRep{} of the \CtlXFiveSeedsN{} seeds replicates by the declared reading. Seeds 44 and 45's feature sets and
constants were recorded in an amendment committed before their runs, and their adapters' tokenizers give seed~42's
tokens and offsets on all 7,307 texts of the receivers and the audit pool. Across the three seeds,
\CtlXFiveAdmMin--\CtlXFiveAdmMax{} of the \CtlXFiveOf{} receivers were admitted.
\begin{figure}[t]
\centering
\includegraphics[width=\linewidth]{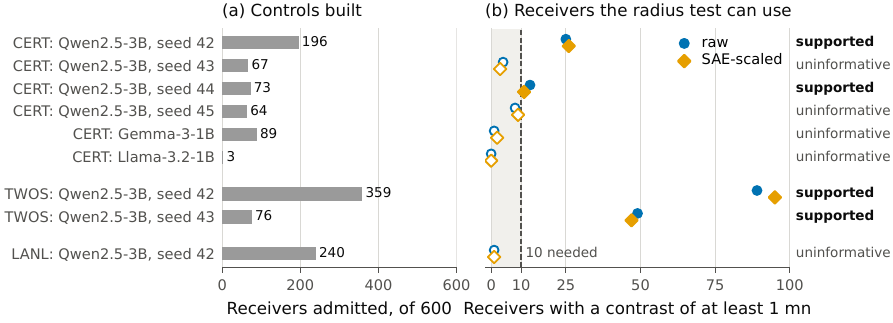}
\caption{Every fine-tuned detector's declared radius test, by training data. (a) Receivers admitted, of the 600 drawn:
those whose removal changed the hidden state and whose controls could be built. (b) Admitted receivers whose contrast
reaches 1\,mn at each standard metric; the test needs at least 10. Filled markers: the test was supported; hollow:
uninformative.}
\label{fig:coverage}
\end{figure}

\subsection{Detectors of two other families}
\label{app:xsix}

A declared replication asks whether the 3B detector's cohort-2 result holds for detectors of two other families,
Llama-3.2-1B and Gemma-3-1B, each fine-tuned from its base model on the same CERT r4.2 data with seed~42. Each
configuration is the 3B's with only the experiment name and the base model changed, which a script checks before the
runs. Llama's adapter was trained by the 3B's command. Gemma's micro-batching differed: 16 sequences per step without
accumulation to step 10,500, then, after an out-of-memory error at step 10,638, eight sequences with two accumulation
steps from the step-10,500 checkpoint to step 17,990, with the data order and random state restored. Its effective batch,
data, order, learning-rate schedule, adapter settings and seed did not change. A check of the resumed run matched its
losses within 0.22\% and its learning rates exactly, but failed its 5\% gradient-norm criterion, which a test showed
the original procedure cannot meet either. Each detector's adaptation-delta SAE (top-$k$ 4, twice as many latents as
the hidden width) and top-five feature set were made by seed~42's commands at about two thirds of its depth (Llama's
layer 11 of 16, Gemma's 17 of 26). Its feature set and constants were recorded in an amendment committed before its
runs, and its adapter's tokenizer gives its base model's tokens and offsets on all 7,307 texts of the receivers and the
audit pool. Each detector then runs cohort 2's two declared tests on the 3B's 600 receiver-days, and is read by the
seeds' replication rule (Appendix~\ref{app:xfive}).

The Gemma detector admitted \CtlXSixGemmaAdmitted{} of the \CtlXFiveOf{} receivers, leaving \CtlXSixGemmaHOneRaw{} and
\CtlXSixGemmaHOneSae{} with contrasts of at least 1\,mn, against the radius test's declared minimum of 10. So that test
was \CtlXSixGemmaXFourVerdict{} and the detector \CtlXSixGemmaVerdict. Of the radius test's other three predictions,
two held at both metrics; the third, a median $r_{0.05}/r^\ast$ of at least 5, failed at the raw metric
(\CtlXSixGemmaKThreeRaw) and held at the SAE-scaled one (\CtlXSixGemmaKThreeSae). Under the first cohort's rules, the
per-receiver test was informative on only \CtlXSixGemmaTInfRaw{} and \CtlXSixGemmaTInfSae{} of \CtlXSixGemmaTOf{}
receivers. The population predictions held at both metrics, on \CtlXSixGemmaPstarRaw{} and \CtlXSixGemmaPstarSae{}
users, but were not informative: the predicted change was \CtlXSixGemmaFalsifRaw{} and \CtlXSixGemmaFalsifSae{} of the
declared four standard errors. The combined rule read \emph{inconsistent} at both metrics, because the per-receiver
$r^\ast$ prediction failed, as at seed~42. The population contrast was negative at both metrics,
\CtlXSixGemmaThetaRaw{} and \CtlXSixGemmaThetaSae\,mn.

The Llama detector admitted only \CtlXSixLlamaAdmitted{} of the \CtlXFiveOf{} receivers; for the other
\CtlXSixLlamaNotConstructed, the control could not be constructed. None of the three reached 1\,mn and only one was
significant at either metric, so every radius prediction was \CtlXSixLlamaXFourVerdict, and its population sample had
\CtlXSixLlamaPsample{} users, below the declared minimum of 25. It therefore did not replicate the 3B's result. Before
its analysis, the self-test of the check that recomputes the predictions failed although every check passed: with one
admitted receiver per user, that self-test's wrong version of the population average coincided with the right one. An
amendment committed before the rerun replaced it, in that case only, with a wrong membership, which the check caught.

\subsection{A classic neural detector: an autoencoder}
\label{app:neural}

\paragraph{Design (declared).}
The detector is the usual unsupervised design for CERT activity: an autoencoder on 26 per-user-day features (session
counts and durations, logon, USB, file, e-mail and web activity, after-hours shares), standardized on the training
split, with architecture 26--64--16--64--26 and the day's mean squared reconstruction error as its score. The site is
the first hidden layer (64 coordinates), with a TopK SAE of 128 latents. Two declared rules select five features each
on cohort-1 days: by first-order attribution to the score, and by mean activation. The confirmation days are cohort 2's
training days, as for the 3B detector and the public models on CERT, and the test is the public models' protocol
unchanged, with one position per day. The contrast's unit is the score's own, not nats.

\paragraph{The declared verdict, and why it could not pass.}
Both rules \emph{do not replicate}. The protocol kept the public models' requirement that the contrast be large against
the rate vector's spread, $G=(\|c\|_1/\|c\|_2)/z\ge5$ with $z=\sqrt{2\ln40}$. Since $\|c\|_1/\|c\|_2\le\sqrt d$, a site
with $d=64$ coordinates has $G\le2.9$, so this requirement could not be met. We found this after the result. Without
it, the activation rule's run would replicate; the attribution rule's would still fail its erasure requirement. With days clustered by
user, at most 20\% of the contrast remains (one-sided 95\%) in \CtlResAeHold{} of the \CtlResAePairs{} rule--metric
pairs; the activation rule reverses it at the SAE-scaled metric, and the attribution rule is not resolved there. Random metrics at the radius move the contrast
by up to \CtlAeBlindMax\%, and the outcome-aware metric lies \CtlAeGapMin--\CtlAeGapMax\% of the way between the two
standard metrics.

\paragraph{The width sweep and the set size (declared predictions; Section~\ref{sec:sensitivity}).}
To test whether the site's dimension is what makes random checks informative here, the same detector was trained at
hidden widths 16, 64, 256 and 1,024 (the SAE keeping two latents per coordinate) and, at width 64, with set sizes
$K_S$ of 1, 3, 5 and 10. Width 16's attribution rule stopped at a declared selection rule (fewer than five candidate
features had positive attribution) and its activation rule at a declared first-order gate; $K_S=1$'s attribution rule
also stopped at that gate. Two predictions were declared before any outcome: the median $d_{\mathrm{eff}}$ rises with
width, and the median largest random change falls with it. Both hold (\CtlWidthLoDeff, \CtlWidthMidDeff,
\CtlWidthHiDeff; \CtlWidthLoRand\%, \CtlWidthMidRand\%, \CtlWidthHiRand\%). A third, that first order predicts the
native outcome-aware change in every pair, fails in \CtlWidthThreeFail{} of \CtlWidthThreePairs{} ($K_S=10$, attribution
rule, SAE-scaled metric, where the radius reaches its cap). The widths differ in more than dimension: capacity, the
SAE's fit and the selected features' rarity also change, so the sweep is an association across widths, not a manipulation of
width alone. At width 64, the activation-based feature rule reversed the contrast at each set size run ($K_S$ of 1, 3, 5
and 10), leaving \CtlKsActMin\% to \CtlKsActMax\% of it (point estimates), while the attribution rule's residual grew with
$K_S$: \CtlKsAttrThreeMin--\CtlKsAttrThreeMax\% at 3, \CtlKsAttrFiveMin--\CtlKsAttrFiveMax\% at 5 and
\CtlKsAttrTenMin--\CtlKsAttrTenMax\% at 10.

\paragraph{On TWOS and LANL (declared before any stage ran on data).}
The same detector, feature rules and test were then declared for the TWOS and LANL windows on which the language-model
detectors were tested (Appendices~\ref{app:logs} and~\ref{app:xseven}), at widths 64 and 1,024. Each window's features
are its activity counts: on TWOS the mouse and keyboard counts of its session lines and its start time
(\CtlNTwoTwosDin{} kept), on LANL the counts of each authentication type, logon type, orientation and failure, the
distinct users and computers, and the time span (\CtlNTwoLanlDin). The bottleneck keeps the ratio of 16 to 26 inputs,
so that TWOS's fewer features are still compressed, and each detector takes at least as many optimizer steps as the CERT
one. Features were selected on the audit pools that selected the language-model detectors' features; the confirmation
documents are the TWOS detectors' 6,077 eligible training windows and the 50,936 training windows of the LANL
detector's 200 users. No arm replicated. On TWOS no test was informative: \CtlNTwoTwosPMin--\CtlNTwoTwosPMax{} windows
entered it, fewer than the 100 the rule needs. At width 64 both rules chose the same five features. At width 1,024 the
attribution rule stopped at its declared selection rule (fewer than five candidates with positive attribution), and
the activation rule met every other requirement, the spread requirement and erasure included. On LANL, width 64 could
not meet the spread requirement ($G \le \CtlNTwoLanlGMax$), as on CERT, and random metrics moved the contrast by up to
\CtlNTwoLanlRandMax\%; at width 1,024 every contrast lay within four standard errors of zero. Of the two declared
predictions, the median $d_{\mathrm{eff}}$ rose with width on both corpora (\CtlNTwoTwosDeffNarrow{} to
\CtlNTwoTwosDeffWide{} on TWOS, \CtlNTwoLanlDeffNarrow{} to \CtlNTwoLanlDeffWide{} on LANL), and the median largest
random change fell with width on LANL (\CtlNTwoLanlRandNarrow\% to \CtlNTwoLanlRandWide\%) but rose on TWOS
(\CtlNTwoTwosRandNarrow\% to \CtlNTwoTwosRandWide\%). The detectors are weak: their eval AUC is \CtlNTwoTwosAucNarrow{}
and \CtlNTwoTwosAucWide{} on TWOS and \CtlNTwoLanlAucNarrow{} and \CtlNTwoLanlAucWide{} on LANL (descriptive).

\subsection{Detectors fine-tuned on real security logs}
\label{app:xseven}

The 3B's tests were declared before any stage on two Qwen2.5-3B adapters fine-tuned earlier in this project on TWOS
windows serialized in CERT's line format (seeds 42 and 43; the 3B's configuration with three epochs instead of one). TWOS's
training split has 12 users. The receivers are 50 training windows per user, drawn by declared seeds from the 6,077 left
after removing the 817 windows of an earlier audit pool, which also trained each run's SAE at layer 24; the users
themselves are the same in both. The prep, stage, jobs, audits and analyses are those of the detectors of two other families (Appendix~\ref{app:xsix}),
with the reading changed only where the data force it: with 12 users the population predictions cannot reach the population test's
minimum of 25, so a seed replicates the 3B's radius result if its radius test is supported.

Seed 42 admitted \CtlXSevenTwosAAdmitted{} receivers (\CtlXSevenTwosANoop{} removals did nothing,
\CtlXSevenTwosANotConstructed{} controls could not be constructed) and seed 43 \CtlXSevenTwosBAdmitted{}
(\CtlXSevenTwosBNoop{} and \CtlXSevenTwosBNotConstructed). Both radius tests were supported (over \CtlXSevenTwosAHOneRaw{} and
\CtlXSevenTwosAHOneSae{} receivers whose contrasts reach 1\,mn in seed 42, and \CtlXSevenTwosBHOneRaw{} and
\CtlXSevenTwosBHOneSae{} in seed 43), the ratio prediction held, and
the predictions that contrasts below 1\,mn, and all contrasts together, would be robust failed wherever informative. The first cohort's
per-receiver test was validated at both metrics (\CtlXSevenTwosATInfRaw{} and \CtlXSevenTwosATInfSae{} of
\CtlXSevenTwosATOf{} receivers informative in seed 42, \CtlXSevenTwosBTInfRaw{} and \CtlXSevenTwosBTInfSae{} of
\CtlXSevenTwosBTOf{} in seed 43), its radius predictions held, and its combined rule read consistent; both independent
checks agree in both seeds.

\paragraph{The LANL detector.}
The same tests were declared before any stage on the 3B adapter fine-tuned earlier in this project on LANL 2015
authentication logs, 32 events per window on one line of text (seed 42, the 3B's configuration): an intrusion detector,
not an insider-threat one. The adapter had been used before for detection and for an identity-ablation experiment on
held-out windows; this test reads only windows that trained it, and no held-out detection. The format has none of CERT's
line types, so the behavior mask covers the whole window and $\theta$ is the contrast in the whole window's loss. The
receivers are three training windows from each of 200 users drawn by declared seeds, cohort 2's design; a new audit pool of
2,432 evaluation windows trained the SAE at layer 24 and selected its features. Three amendments, each committed before the step it changed
and none after a result, fixed the port: the first gave the prep's input CERT's session counts as empty fields, which
the extractor only copies; the second set the features, the constants and 24 construction workers for LANL's longer
windows, which change only the wall time; the third split the native scoring over two GPUs when its projected end
passed the declared cutoff, and the receiver scored on both was identical.

\CtlXEightAdmitted{} of the 600 receivers were admitted, from \CtlXEightAdmittedUsers{} users
(\CtlXEightNotConstructed{} controls could not be constructed, \CtlXEightNoop{} removals did nothing). The radius test was
\CtlXEightXFourVerdict: of the \CtlXEightSignifRaw{} and \CtlXEightSignifSae{} receivers whose contrasts were
significant, one reached 1\,mn at each metric, against the 10 required. The prediction that contrasts below 1\,mn would
be robust failed (median $r^\ast$ \CtlXEightKOneRaw{} and \CtlXEightKOneSae{}, under the cap of \CtlXEightRCap), and the
predictions that all contrasts together would be robust and the ratio prediction held. Since the declared reading needs
the radius test supported, the 3B's result does not replicate on LANL whatever the population test shows.
The population test was informative at both metrics (\CtlXEightPstarRaw{} and \CtlXEightPstarSae{} users) and its
predictions held at both metrics with its users fixed, but at the raw metric the predicted change was
\CtlXEightFalsifRaw{} of the declared four standard errors, so that prediction was not informative. The outcome-aware metric at $r^\ast$ reversed
the population contrast natively, leaving \CtlXEightLeftRaw\% and \CtlXEightLeftSae\% of it, while random metrics
moved it by up to \CtlXEightRandomRaw\% and \CtlXEightRandomSae\%; but the contrast lies only \CtlXEightThetaTRaw{}
and \CtlXEightThetaTSae{} standard errors from zero across users, so none of these shares is resolved. The
per-receiver test was validated at both metrics (\CtlXEightTInfRaw{} and \CtlXEightTInfSae{} of \CtlXEightTOf{}
receivers informative), the ratio prediction held, and the prediction that the median receiver's radius is at most 0.5
failed, so the combined rule reads inconsistent at both metrics, as for the 3B's first cohort; both independent
checks agree. Under the third amendment, the receiver scored on both GPUs was identical in both records.

\FloatBarrier\section{Every run of the test}
\label{app:runs}

Table~\ref{tab:runs} gives every run that Table~\ref{tab:settings} summarizes, each at the raw / SAE-scaled metric; the
six web-text replication blocks (Appendix~\ref{app:public}) are in neither table. Its columns: $n$, the units in the population contrast
(users for the detectors, documents or days otherwise); $\theta$, the declared population contrast, in mn ($10^{-3}$
nats; for the autoencoder, $10^{-3}$ of its score); $r^\ast$/gap, the outcome-aware metric's Hilbert distance from the
declared one, as a percentage of the distance between the two standard metrics; cost, $e^{r^\ast}$, the largest change
that metric makes to the relative cost of two edits; the residual, the contrast left under it as a percentage of
$\theta$ (0 erased, below 0 reversed, $-$0 just below 0), with its one-sided 95\% bound (post hoc; $\infty$: $\theta$ not
distinguishable from zero across users); random, the larger change from two random metrics at the same distance, as a
percentage of $\theta$; and the verdict, the declared reading, followed on the security logs by how many of the run's
two replications on fresh documents replicated. The Qwen3-8B run is exploratory and Pythia's float64 rerun is post hoc.

The runs are \CtlSetPrimary{} declared primary runs, \CtlSetReplications{} replications of the security-log runs on
fresh documents, \CtlXFiveSeedsN{} further training seeds of the 3B detector, a Gemma-3-1B detector, a 3B detector
fine-tuned on LANL, the exploratory Qwen3-8B run and Pythia's post hoc float64 rerun. Four further declared
language-model runs contribute no pairs: Pythia's float32 run on web text, which stopped at its first-order gate; a
Llama-3.2-1B detector that admitted too few receivers (days whose controls could be constructed) for either test; and two
3B detectors fine-tuned on TWOS, read by their radius test alone. Table~\ref{tab:settings} counts these four among its
runs.

\begin{table}[t]
\centering
{\footnotesize\setlength{\tabcolsep}{3pt}\begin{tabular}{@{}lr@{\,/\,}rr@{\,/\,}rr@{\,/\,}rr@{\,/\,}rr@{\,/\,}rr@{\,/\,}rrl@{}}
\toprule
 & \multicolumn{2}{c}{} & \multicolumn{2}{c}{} & \multicolumn{4}{c}{Size of the change} & \multicolumn{4}{c}{Residual (\% of $\theta$)} &  &  \\
\cmidrule(lr){6-9}\cmidrule(lr){10-13}
Run & \multicolumn{2}{c}{$n$} & \multicolumn{2}{c}{$\theta$ (mn)} & \multicolumn{2}{c}{$r^\ast$/gap (\%)} & \multicolumn{2}{c}{cost ($\times$)} & \multicolumn{2}{c}{estimate} & \multicolumn{2}{c}{95\% bound} & Random (\%) & Verdict \\
\midrule
\multicolumn{15}{@{}l}{\textit{CERT r4.2 detectors (QLoRA adapters, adaptation-delta SAEs)}} \\
Qwen3-8B & 32 & 33 & 4.02 & 4.14 & 2.0 & 1.8 & 1.22 & 1.20 & $-$36 & $-$37 & $-$21 & $-$10 & 2.8 & exploratory \\
Qwen2.5-3B, cohort 1 & 61 & 62 & 0.55 & 0.43 & 6.6 & 4.9 & 1.40 & 1.28 & $-$18 & $-$37 & 52 & 67 & 8.8 & inconsistent$^{a}$ \\
Qwen2.5-3B, cohort 2 & 77 & 75 & 0.51 & 0.32 & 5.9 & 4.8 & 1.35 & 1.27 & 3 & 32 & 33 & 73 & 7.4 & supported$^{b}$ \\
Qwen2.5-3B, cohort 2, seed 43 & 31 & 30 & 0.15 & 0.35 & 15.0 & 12.3 & 2.15 & 1.87 & 31 & $-$6 & 373 & $\infty$ & 17.4 & does not replicate$^{c}$ \\
Qwen2.5-3B, cohort 2, seed 44 & 26 & 27 & 0.22 & 0.73 & 3.2 & 2.4 & 1.17 & 1.13 & 18 & 37 & 123 & $\infty$ & 6.0 & replicates \\
Qwen2.5-3B, cohort 2, seed 45 & 28 & 25 & 0.29 & 0.25 & 2.1 & 2.3 & 1.11 & 1.12 & 42 & 26 & $\infty$ & 96 & 12.0 & does not replicate$^{c}$ \\
Gemma-3-1B, cohort 2 & 43 & 42 & $-$0.10 & $-$0.10 & 74.0 & 93.1 & 8{,}992 & 93{,}999 & 96 & 39 & 132 & 78 & 18.6 & does not replicate$^{c}$ \\
Llama-3.2-1B, cohort 2 & \multicolumn{13}{l}{3 of 600 receivers admitted; too few for either test} & does not replicate$^{c}$ \\
\addlinespace
\multicolumn{15}{@{}l}{\textit{CERT r4.2 days read by public models (public SAEs; days clustered by user)}} \\
Llama-3.2-1B & \multicolumn{2}{c}{998} & 3.59 & 2.73 & 3.0 & 2.2 & 1.16 & 1.12 & 0 & 0 & 2 & 2 & 6.6 & replicates \\
Gemma-3-1B & \multicolumn{2}{c}{1000} & 17.77 & 13.65 & 5.3 & 4.1 & 1.70 & 1.51 & $-$1 & $-$3 & $-$0 & $-$2 & 6.9 & replicates \\
Pythia-410m, float64 & 963 & 964 & 5.15 & 4.31 & 5.2 & 4.4 & 1.16 & 1.14 & 0 & 0 & 1 & 1 & 3.9 & replicates \\
\addlinespace
\multicolumn{15}{@{}l}{\textit{CERT r4.2 days, an autoencoder detector ($\theta$ in $10^{-3}$ of its score)$^{d}$}} \\
Autoencoder, attribution & 748 & 761 & 546.1 & 579.0 & 48.3 & 73.0 & 4.92 & 11.08 & 11 & 19 & 17 & 23 & 34.7 & does not replicate$^{d}$ \\
Autoencoder, activation & 1163 & 1170 & 1182.3 & 1412.8 & 16.6 & 21.0 & 1.73 & 2.00 & $-$18 & $-$35 & 5 & $-$16 & 17.8 & does not replicate$^{d}$ \\
\addlinespace
\multicolumn{15}{@{}l}{\textit{TWOS and LANL windows, the same autoencoder (Appendix~\ref{app:neural})}} \\
Autoencoder, TWOS, width 64 & \multicolumn{13}{l}{both rules chose the same five features; 75--76 windows, under the 100 needed} & uninformative \\
Autoencoder, TWOS, width 1,024 & \multicolumn{13}{l}{activation rule: 93 windows, under 100; attribution rule stopped at selection} & uninformative \\
Autoencoder, LANL, width 64 & \multicolumn{13}{l}{spread requirement cannot pass ($G \le 2.1$), as in note $d$} & does not replicate \\
Autoencoder, LANL, width 1,024 & \multicolumn{13}{l}{every contrast within four standard errors of zero} & uninformative \\
\addlinespace
\multicolumn{15}{@{}l}{\textit{OpenWebText (public models, public SAEs)}} \\
GPT-2 small & \multicolumn{2}{c}{332} & 1.37 & 1.27 & 17.6 & 15.8 & 2.69 & 2.43 & $-$7 & $-$8 & 15 & 16 & 9.6 & replicates \\
DeepSeek-R1-Distill-Qwen-1.5B & 886 & 883 & 1.05 & 1.21 & 11.9 & 12.6 & 1.85 & 1.92 & 4 & 5 & 11 & 11 & 6.3 & replicates \\
Gemma-3-1B & \multicolumn{2}{c}{883} & 2.75 & 2.86 & 11.5 & 16.9 & 3.60 & 6.54 & $-$65 & $-$42 & $-$46 & $-$24 & 10.0 & replicates \\
Llama-3.2-1B & 678 & 679 & 0.56 & 0.59 & 11.2 & 11.6 & 1.77 & 1.80 & $-$5 & $-$5 & 27 & 26 & 9.5 & replicates \\
Pythia-410m, float32 (declared) & \multicolumn{13}{l}{stopped at its first-order gate (5 of 6 cases passed)} & does not replicate \\
Pythia-410m, float64 (post hoc) & \multicolumn{2}{c}{928} & 0.20 & 0.47 & 2.0 & 5.9 & 1.06 & 1.17 & $-$7 & $-$22 & 14 & $-$11 & 3.9 & replicates \\
\addlinespace
\multicolumn{15}{@{}l}{\textit{TWOS and LANL security logs (public models; verdict, then replications on fresh documents)}} \\
Llama, TWOS & 957 & 959 & 3.26 & 2.24 & 5.4 & 3.4 & 1.30 & 1.18 & $-$1 & $-$1 & 8 & 12 & 4.1 & replicates; 1 of 2 \\
Gemma, TWOS & \multicolumn{2}{c}{1000} & 10.37 & 12.53 & 2.0 & 2.9 & 1.23 & 1.35 & 0 & 0 & 6 & 3 & 7.1 & replicates; 2 of 2 \\
Pythia, TWOS & \multicolumn{2}{c}{978} & 6.96 & 6.22 & 6.3 & 5.7 & 1.21 & 1.19 & 0 & 0 & 2 & 3 & 5.3 & replicates; 2 of 2 \\
Llama, LANL & \multicolumn{2}{c}{999} & 1.31 & 1.18 & 1.8 & 1.6 & 1.09 & 1.09 & 0 & 0 & 4 & 4 & 3.7 & replicates; 2 of 2 \\
Gemma, LANL & \multicolumn{2}{c}{1000} & 7.08 & 8.32 & 4.1 & 5.6 & 1.47 & 1.69 & 1 & $-$1 & 4 & 1 & 5.5 & replicates; 1 of 2 \\
Pythia, LANL & \multicolumn{2}{c}{950} & 2.20 & 2.15 & 6.6 & 6.4 & 1.27 & 1.26 & $-$1 & $-$1 & 1 & 1 & 7.2 & replicates; 2 of 2 \\
\addlinespace
\multicolumn{15}{@{}l}{\textit{Detectors fine-tuned on TWOS and LANL (QLoRA adapters, adaptation-delta SAEs)}} \\
Qwen2.5-3B, TWOS, seed 42 & \multicolumn{13}{l}{359 of 600 admitted; radius test supported, median $r^\ast$ 0.27 / 0.28} & replicates$^{e}$ \\
Qwen2.5-3B, TWOS, seed 43 & \multicolumn{13}{l}{76 of 600 admitted; radius test supported, median $r^\ast$ 0.08 / 0.08} & replicates$^{e}$ \\
Qwen2.5-3B, LANL, seed 42 & 143 & 145 & $-$0.02 & $-$0.01 & 7.8 & 9.7 & 1.45 & 1.58 & $-$9 & $-$8 & $\infty$ & $\infty$ & 27.5 & does not replicate$^{c}$ \\
\bottomrule
\end{tabular}}
\caption{Every run of the test. $^{a}$Population predictions held; the per-receiver prediction (median
$r^\ast \le 0.5$) failed. $^{b}$A radius test, on receivers with contrasts of at least 1\,mn, of a claim formed after
cohort 1's result; it held, and under cohort 1's rule the run reads as $^{a}$. $^{c}$The radius test was uninformative
(too few receivers); replicating needs it supported. $^{d}$A site of 64 coordinates allows
$G \le \sqrt{64}/2.72 = 2.9$, below the required 5; the attribution rule also failed erasure. $^{e}$TWOS has 12 users,
under the 25 the population test needs, so the reading is the radius test's; median per-receiver $r^\ast$, raw /
SAE-scaled.}
\label{tab:runs}
\end{table}
\end{document}